\documentclass[12pt]{article}

\usepackage[table]{xcolor}
\usepackage{graphicx}
\usepackage{amsmath}
\usepackage{float}
\usepackage{comment}
\usepackage{amsthm}
\usepackage{amssymb}
\usepackage{mathtools}
\usepackage{amsfonts}
\usepackage{mathrsfs}
\usepackage{tikz-cd}
\usepackage{booktabs}
\usepackage{caption}
\usepackage[colorinlistoftodos]{todonotes}

\usepackage[natbibapa]{apacite}   

\newtheorem{theorem}{Theorem}
\newtheorem{lemma}{Lemma}

\theoremstyle{definition}
\newtheorem{definition}{Definition}

\theoremstyle{remark}
\newtheorem{remark}{Remark}

\usepackage{hyperref}   
\title{Highly Adaptive Principal Component Regression}

\author{
Mingxun Wang$^{1}$, Alejandro Schuler$^{1}$,\\
Mark van der Laan$^{1}$, Carlos García Meixide$^{1,2}$\\[0.5cm] 
$^{1}$University of California, Berkeley\\
$^{2}$ICMAT, National Research Council of Spain
}

\date{}

\begin{document}

\maketitle

\begin{abstract}
    The Highly Adaptive Lasso (HAL) is a nonparametric regression method that
achieves almost dimension-free convergence rates under minimal smoothness
assumptions, but its implementation can be computationally prohibitive in high
dimensions due to the large design matrix it requires. The Highly Adaptive
Ridge (HAR) has been proposed as a related ridge-regularized analogue. Building
on both procedures, we introduce the Principal Component Highly Adaptive Lasso
(PCHAL) and Principal Component Highly Adaptive Ridge (PCHAR). These estimators
use an outcome-blind principal-component reduction of the HAL basis, offering
substantial computational gains over HAL while achieving empirical performance
comparable to HAL and HAR. We also describe an early-stopped gradient descent
variant, which provides a convenient form of smooth spectral regularization
without explicitly selecting a hard principal-component cutoff. Finally, we
uncover that under special circumstances, the HAL kernel is identical to the
covariance function of Brownian motion.
\end{abstract}

\section{Introduction}

Adaptive nonparametric regression---methods that automatically adjust to the unknown complexity of the target function---is a cornerstone of modern statistical learning.
Classical theory characterizes attainable risk rates under smoothness or structural constraints \citep{stone1982optimal}, while modern practice relies on estimators that combine rich function classes with regularization and data-driven model selection \citep{donoho1994ideal}.
A prominent and historically influential route to adaptivity is through spline-based expansions and penalization, including smoothing splines \citep{wahba1990spline,green1994nonparametric}, penalized B-splines (P-splines) \citep{eilers1996flexible}, and locally adaptive regression splines \citep{mammen1997locally}.
Related adaptive basis-selection ideas also appear in multivariate adaptive regression splines (MARS) \citep{friedman1991mars} and tree-based step-function models such as CART \citep{breiman1984cart}.
Stepwise–constant (lower–orthant indicator) basis functions have a long history
in nonparametric regression. Early work by \citep{wong1995} established
convergence rates for sieve estimators built from partition-based function
classes. Histogram and partitioning estimators were further developed
systematically in \citep{gyorfi2002distribution}. Later,
\citep{buhlmann2003boosting} showed that boosting with step functions can yield
adaptive estimators in both regression and classification.

In parallel, sparsity-inducing regularization---especially the Lasso \citep{tibshirani1996regression}---has become a standard mechanism for adaptive selection among a large dictionary of basis functions, with scalable coordinate-descent implementations enabling high-dimensional use in practice \citep{friedman2010glmnet}.

Closely connected ``$\ell_1$-type'' nonparametric estimators include total-variation regularization \citep{rudin1992tv}, the fused lasso \citep{tibshirani2005fused}, and trend filtering \citep{kim2009trend,tibshirani2014trend}, all of which highlight the interplay between rich piecewise-polynomial/step-function representations and convex regularization.

The \emph{Highly Adaptive Lasso} (HAL) \citep{benkeser2016highly,van2017generally,van2023higher,fang2021hk} is a more recent method for nonparametric regression that has attracted some attention in the literature.
HAL constructs a very large linear span of indicator (or, more generally, spline-like) basis functions indexed by axis-aligned knots, and fits the regression via an $\ell_1$-penalized empirical risk minimization \citep{tibshirani1996regression}.
This construction is closely related to the older saturated-spline perspective in which a sufficiently rich basis can interpolate complex shapes while regularization controls complexity \citep{wahba1990spline,green1994nonparametric,mammen1997locally,eilers1996flexible}.

A key theoretical feature of HAL is that the $\ell_1$ constraint/penalty on the coefficients corresponds to a bounded \emph{sectional variation norm} (a multivariate analogue of bounded variation), which provides an interpretable, low-smoothness complexity measure for high-dimensional regression functions \citep{benkeser2016highly,van2023higher}.
When the true regression function has bounded sectional variation norm, the zero-order HAL estimator achieves the convergence rate
\[
O_P\!\left(n^{-1/3}(\log n)^{2(d-1)/3}\right),
\]
where the ambient dimension $d$ enters only through a logarithmic factor \citep{bibaut2019fast,fang2021hk}.

Because of its rate guarantees, HAL has become a convenient nuisance-function learner for
semiparametric inference, where the goal is valid estimation and uncertainty quantification
for a low-dimensional target parameter in the presence of infinite-dimensional nuisance components.
Classical efficiency theory characterizes optimal procedures through tangent spaces and efficient
influence functions (EIFs) \citep{bickel1993efficient,vdvaart1998asymptotic,tsiatis2006semiparametric}.
In missing-data and causal-inference settings, foundational work by Robins and collaborators developed
inverse-probability-weighted estimating equations and semiparametric efficient procedures, including
early doubly-robust constructions \citep{robins1994regressors,robins1995efficiency,robins1995repeated}.
Targeted Maximum Likelihood Estimation (TMLE) can be viewed as a likelihood-based, plug-in approach
that performs a targeted fluctuation step so that the resulting estimator solves (approximately) the EIF
estimating equation while allowing flexible machine learning for nuisance estimation \citep{van2006tmle,vdlaan2011targeted,van2017generally}.
More recently, Double/Debiased Machine Learning (DML) formalizes closely related principles---Neyman-orthogonal
scores and cross-fitting---to enable $\sqrt{n}$-valid inference when nuisance functions are estimated by
high-dimensional or nonparametric learners \citep{chernozhukov2018dml}. All of these methods typically rely on functional nuisance estimates that converge at fast-enough rates in order to remove first-order bias.

Despite these appealing statistical properties, the practical deployment of HAL can be severely limited by computational cost.
In its most direct implementation, the HAL design matrix has dimension $n \times p$, where $p$ equals the number of candidate basis functions and can be as large as $p=n(2^d-1)$ for $d$ covariates.
Thus, even when fitting is performed with efficient convex optimization methods for the Lasso \citep{tibshirani1996regression,friedman2010glmnet}, both (i) constructing the basis/design and (ii) repeatedly solving high-dimensional $\ell_1$-penalized problems (e.g., for cross-validation) can become prohibitive.
These difficulties motivate algorithmic reformulations that avoid working directly in the full basis space.

A recent step in this direction is the \emph{Highly Adaptive Ridge} (HAR) estimator \citep{schuler2024highlyadaptiveridge}, which replaces the $\ell_1$ penalty in HAL with an $\ell_2$ penalty \citep{hoerl1970ridge}. Under slightly stricter assumptions, HAR achieves the same convergence rate as HAL.
This change enables a kernelized representation which will be elaborated later: writing the (implicit) HAL feature map as $H\in\mathbb{R}^{n\times p}$ and the associated kernel matrix as $K = HH^\top\in\mathbb{R}^{n\times n}$, the fitted values can be expressed in closed form via the Sherman--Morrison--Woodbury identity \citep{sherman1950morrison,woodbury1950matrix}.
This viewpoint connects HAR to classical kernel ridge regression and the broader literature on kernel methods \citep{scholkopf2002learning,hastie2009elements}, while retaining HAL's data-adaptive feature construction through knots.
From a computational perspective, working with $K$ shifts the bottleneck from a potentially enormous $p$ to an $n\times n$ matrix, which is often advantageous when $p\gg n$.
\paragraph{Contribution.}
In this work, we propose two new estimators---\emph{Principal Components Highly Adaptive Lasso} (PCHAL) and \emph{Principal Components Highly Adaptive Ridge} (PCHAR)---that reduce the computational cost of HAL/HAR while aiming to preserve predictive performance.
Our key idea is to approximate the HAL/HAR kernel matrix $K$ by its leading $k$ principal components, yielding a low-rank representation that projects the regression problem into a $k$-dimensional orthogonal score space.
Low-rank kernel approximations and spectral truncation are classical tools in kernel learning (e.g., kernel PCA and Nystr\"om-type approximations) \citep{scholkopf1998kpca,williams2001nystrom}, and scalable randomized algorithms for truncated eigendecompositions/SVD are well developed \citep{halko2011randomized}.
Our contribution is to bring this spectral perspective \emph{specifically to the HAL/HAR kernel induced by saturated, knot-based features}, and to highlight an \emph{outcome-blind} structure: the embedding is determined entirely by the covariates through their relative positions, not by the response $Y$.
In particular, the eigen-score map can be viewed as a universal, geometry-driven reparameterization of the sample---once the covariates are placed (or sorted, in settings where such a representation applies), the resulting eigen-basis is fixed up to this relative positioning.
Exploiting this structure yields estimators with simple analytic forms in the truncated score domain:
\begin{itemize}
\item \textbf{PCHAR:} an $\ell_2$-penalized estimator that inherits the closed-form ridge solution in the $k$-dimensional orthogonal basis.
\item \textbf{PCHAL:} an $\ell_1$-penalized estimator that, due to orthogonality of the retained scores, reduces to \emph{componentwise soft-thresholding} (a direct analogue of the orthogonal-design Lasso solution) \citep{tibshirani1996regression}.
\end{itemize}
These closed-form solutions eliminate iterative optimization in model fitting and substantially accelerate cross-validation by reducing each refit to (i) a truncated spectral computation and (ii) closed-form solutions in $\mathbb{R}^k$.

Ready-to-use R and Python implementations, including cross-validation routines, are publicly available at \url{https://github.com/meixide/hapc}.

\section{Highly Adaptive Lasso and Ridge}
In this section, we review the Highly Adaptive Lasso and Highly Adaptive Ridge estimators. Many of the definitions introduced below also appear in \citep{owen2005multidimensional,benkeser2016highly,schuler2024highlyadaptiveridge,fang2021hk}. For notational and technical convenience, we formulate the results on $[0,1]^d$. More generally, the same framework applies to any compact axis-aligned rectangle after an affine rescaling to $[0,1]^d$; in particular, in one dimension one may replace $[0,1]$ by $[0,\tau]$. The main goal of this section is to introduce the sectional representation theorem, Theorem~\ref{thm:sectional_rep}. Readers who are primarily interested in the estimator may consult \citep{benkeser2016highly,schuler2024highlyadaptiveridge} for complete explanations. At a high level, the key idea behind HAL and HAR is that the value of a function at a given point admits the linear representation in Equation~\ref{eq:discrete_rep}, which makes it possible to sidestep many of the underlying measure-theoretic subtleties. Readers who are willing to take the sectional representation theorem for granted may therefore begin from Equation~\ref{eq:sectional_rep}. 

For a nonempty index set \(s \subseteq \{1,\ldots,d\}\) and a vector \(x=(x_1,\ldots,x_d)\in[0,1]^d\),
we write
\[
x_s := (x_j)_{j\in s}\in[0,1]^{|s|},
\qquad
0_s := (0,\ldots,0)\in[0,1]^{|s|}.
\]
Given \(f:[0,1]^d\to\mathbb R\), define the \(s\)-section (anchored at \(0\)) by
\[
f_s:[0,1]^{|s|}\to\mathbb R,
\qquad
f_s(u_s) := f(u_s,0_{-s}),
\]
where \((u_s,0_{-s})\in[0,1]^d\) denotes the vector whose coordinates in \(s\) equal \(u_s\) and whose
coordinates in the complement \(-s:=\{1,\ldots,d\}\setminus s\) are zero.
We also use the rectangle notation
\[
(0_s,x_s] := \prod_{j\in s}(0,x_j] \subset (0,1]^{|s|}.
\]

\begin{definition}
    Let $f(x)$ be a function on $[0,1]^d$. Let $a=(a_1,\ldots,a_d)$ and $b=(b_1,\ldots,b_d)$ be elements of $[0,1]^d$ such that $a_i<b_i$ for all $i$. A vertex of $[a,b]$ is of the form $(c_1, \dots, c_d)$ where each $c_i \in \{a_i, b_i\}$. Let $\mathcal{V}([a,b])$ be the set of all $2^d$ vertices of $[a,b]$. We define the sign of a vertex $v = (c_1, \dots, c_d) \in \mathcal{V}([a,b])$ as
\[
\text{sign}(v) \equiv (-1)^{\sum_{i=1}^d 1(c_i = a_i)}.
\]
That is, the sign of the vertex $v$ is $-1$ if $v$ contains an odd number of $a_i$, and $1$ if it contains an even number of $a_i$.
\end{definition} 

\vspace{1em}
\begin{definition}
Let $D([0,1]^d)$ denote the class of functions $f:[0,1]^d\to\mathbb R$ that are
right--continuous with existing left limits in each coordinate (multivariate c\`adl\`ag) \citep{neuhaus1971weak}.
\end{definition}

\begin{definition}
    The generalized difference of $f \in D([0,1]^d)$ over an axis-parallel box $A=[a,b] \subset (0,1]^d$ with vertices $\mathcal{V}(A)$ is defined as
\[
\Delta^{(d)}(f;A) \equiv \sum_{v \in \mathcal{V}([a,b])} \text{sign}(v) f(v).
\]
\end{definition} 

For $s=1,\ldots,d$, let 
\[
0 = x_0^{(s)}<x_1^{(s)}<\cdots< x_{m_s}^{(s)}=1
\]
be a partition of $[0,1]$, and let $\mathcal{P}$ be the partition of $[0,1]^d$ which is given by 
\[
\mathcal{P}= \left\{\left[x_{l_1}^{(1)},x_{l_1+1}^{(1)}\right]\times\cdots\times\left[x_{l_d}^{(d)},x_{l_d+1}^{(d)}\right]: l_s = 0,\ldots, m_s-1,s=1,\ldots,d\right\}
\]
\begin{definition}
    The variation of $f$ on $[0,1]^d$ in the sense of Vitali is given by 
    \[
    V^{(d)}(f;[0,1]^d)=\sup_{\mathcal{P}}\sum_{A\in\mathcal{P}}|\Delta^{(d)}(f;A)|
    \]
    where the supremum is extended over all partitions of $[0,1]^d$ into axis-parallel boxes generated by $d$ one-dimensional partitions of $[0,1]$.
\end{definition}

\begin{definition}
    The variation of $f$ on $[0,1]^d$ in the sense of Hardy and Krause (also called \textit{sectional variation}) is given by 
    \[
    \text{Var}_{HK0}(f;[0,1]^d)=\sum_{\emptyset \neq s\subseteq\{1,\ldots,d\}} V^{(|s|)}(f_s;[0,1]^s)
    \] 
\end{definition}
\begin{remark}
    The Hardy--Krause variation anchored at 0 is almost identical to the sectional variation except for the term $f(0)$. We write $V(f)$ to replace $\text{Var}_{HK0}(f;[0,1]^d)$ to ease the notation.
\end{remark}

The following theorem is a key analytical tool underlying the HAL framework.
For completeness, a proof is given in the appendix; see
\citep{gill1995inefficient,benkeser2016highly,van2018targeted}.
\begin{theorem}[Sectional Representation of C\`adl\`ag Functions]\label{thm:sectional_rep}
Let \( f: [0,1]^d \to \mathbb{R} \) be a c\`adl\`ag function with bounded Hardy--Krause variation anchored at 0, i.e., \( V(f) < \infty \). Then \( f \) admits the representation
\[
f(x) = f(0) + \sum_{\emptyset\neq s \subseteq \{1,\ldots,d\}} 
\int_{(0_s,x_s]} \mu_{f_s}(du_s).
\]
\end{theorem}
The proof and the construction of the measures $\mu_{f_s}$ are explained in Appendix~\ref{app:sectional-representation}.
\par Suppose we observe \(n\) i.i.d.\ copies \(O_i=(X_i,Y_i)\sim P_0\), where
\(X_i\in \mathcal X\subset [0,1]^d\) and \(Y_i\in\mathbb R\). Let
\(\mathcal F\) be a prespecified class of real-valued functions on
\(\mathcal X\), encoding the desired regularity constraints (e.g., bounded
sectional variation norm). We define the target regression function \(f_0\) as
the population least-squares minimizer over \(\mathcal F\):
\[
f_0 \in \arg\min_{f\in\mathcal F} P_0\bigl(Y-f(X)\bigr)^2
=
\arg\min_{f\in\mathcal F} \mathbb E_{P_0}\!\left[(Y-f(X))^2\right].
\]
Accordingly, the parameter of interest may be written as
\[
\Psi(P_0) \;=\; f_0 \in \mathcal F \subset \{ f:[0,1]^d\to\mathbb R \}.
\]

The representation in Theorem~\ref{thm:sectional_rep} can be
written as
\begin{equation} \label{eq:sectional_rep}
  f(x) 
  = f(0) 
    + \sum_{\emptyset \neq s \subseteq \{1,\ldots,d\}} 
      \int_{0_s}^{1_s} I(u_s \leq x_s) \, d\mu_{f,s}(u_s).
\end{equation}
\medskip
suggesting an underlying regression structure.
To connect it to a finite-dimensional estimation problem, we approximate each measure
\(\mu_{f,s}\) by a discrete signed measure supported on the observed sample.
For every nonempty subset \(s \subseteq \{1,\ldots,d\}\), let
\(\tilde x_{i,s} := (X_i)_s\) denote the projected covariates, where the tilde indicates that these projected covariates are used as knot points, and define
\[
  \mu_{n,s} \;=\; \sum_{i=1}^n \beta_{s,i}\,\delta_{\tilde x_{i,s}},
\]
for coefficients \(\beta_{s,i}\in\mathbb R\). Substituting \(\mu_{n,s}\) into
\eqref{eq:sectional_rep} yields the finite expansion
\begin{equation}\label{eq:discrete_rep}
  f_{n,\beta}(x)
  \;=\; f(0)
  \;+\; \sum_{\emptyset \neq s \subseteq \{1,\ldots,d\}} \sum_{i=1}^n
  \beta_{s,i}\, I(\tilde x_{i,s} \le x_s).
\end{equation}

Defining basis functions
\[
  \phi_{s,i}(x) := I(x_s \geq \tilde{x}_{i,s}),
\]
we can rewrite~\eqref{eq:discrete_rep} as a linear combination of these basis
functions. In particular, if we set \(\beta_0 := f(0)\), then
\begin{equation}\label{trueclass}
  f_{n,\beta}(x) = \beta_0 + \sum_{\emptyset \neq s \subseteq \{1,\ldots,d\}} 
                 \sum_{i=1}^n \beta_{s,i}\,\phi_{s,i}(x).
\end{equation}
The sectional variation norm of \(f_n\) is then given by
\[
  \|f_{n,\beta}\|_v 
  = |\beta_0| + \sum_{\emptyset \neq s \subseteq \{1,\ldots,d\}} 
                 \sum_{i=1}^n |\beta_{s,i}|,
\]
which mirrors an \(\ell_1\)-norm on the coefficient vector \(\beta\).

For estimation, we restrict attention to functions of the form \(f_{n,\beta}\) and
impose an upper bound on the sectional variation. For a given \(M > 0\), define
the function classes
\[
  \mathcal{F}_{n,M} 
  := \left\{ 
       f_{n,\beta} : 
       |\beta_0| + \sum_{\emptyset \neq s \subseteq \{1,\ldots,d\}} 
                    \sum_{i=1}^n |\beta_{s,i}| \le M 
     \right\}.
\]
We use the empirical risk associated with squared error loss for simplicity
\[
\mathcal{R}_n(f)
\;:=\;
\frac{1}{n}\sum_{i=1}^n \bigl(Y_i - f(X_i)\bigr)^2 .
\]
A \emph{HAL estimator} is any minimizer over the sieve:
\[
\widehat{\beta}_{n}\in\arg\min_{\beta:\, f_{n,\beta}\in\mathcal{F}_{n,M}} \mathcal{R}_n(f),
\qquad
\widehat f_{n}(x)\;:=\;f_{\widehat{\beta}_{n}}(x).
\]
This optimization problem reduces to a standard penalized regression problem. Although we define HAL through the constrained sieve $\mathcal F_{n,M}$, in
practice one works with the equivalent Lagrangian form, replacing the
hard sectional-variation bound by an $\ell_1$ penalty with tuning parameter
$\lambda$. Thus HAL is typically fit as a lasso problem, with $\lambda$
selected by cross-validation.\\

Let \(p := n(2^d - 1)\) denote the total number of non-intercept basis
functions, corresponding to the collection
\(\{\phi_{s,i} : \emptyset \neq s \subseteq \{1,\ldots,d\},\ i=1,\dots,n\}\).
We define the empirical \emph{HAL design matrix}
\(H \in \{0,1\}^{n \times p}\) by
\[
  H_{j,\ell} := \phi_{s,i}(X_j),
\]
where the column index \(\ell \in \{1,\ldots,p\}\) corresponds to a particular
pair \((s,i)\). Stacking the coefficients \(\beta_{s,i}\) into a vector
\(\beta \in \mathbb{R}^p\), the empirical risk can be viewed as a function of
\(\beta\) and the penalty \(\|f_\beta\|_v\) becomes an \(\ell_1\)-type norm on
\((\beta_0, \beta)\).\\

\emph{Highly Adaptive Ridge} (HAR) is the analogue of HAL obtained by replacing the
\(\ell_1\)-type variation constraint with an \(\ell_2\) (ridge) penalty on the coefficients.
Concretely, HAR uses the same HAL dictionary \(\{\phi_{s,i}\}\) and design matrix \(H\), and
estimates \((\beta_0,\beta)\) by minimizing the empirical mean squared error plus a ridge
penalty on \(\beta\), yielding a ridge-regularized version of the HAL expansion. Unlike HAL, an \(\ell_2\) penalty does not directly control the Hardy--Krause/sectional
variation norm (which corresponds to an \(\ell_1\) norm of the coefficients in this saturated
basis). Consequently, rate guarantees for HAR are typically derived under stronger
conditions ensuring effective variation control, e.g., by restricting to (or showing that the
oracle approximation lies in) a \emph{shrinking} \(\ell_2\)-ball such as
\(\|\beta\|_2 \le M n^{-1}\), so that \(\|\beta\|_1\) (and hence the sectional variation) remains
bounded via Cauchy--Schwarz; see, e.g., the appendix of \citep{schuler2024highlyadaptiveridge}.

\section{The Highly Adaptive Kernel Trick}
Now that we have outlined the basic HAL and HAR algorithms, we turn to a discussion of how exactly HAR achieves computational feasibility. 

Concretely, we define the HAL feature map $h(\cdot)$ to be the design vector obtained by evaluating
at a point $x\in[0,\tau]^d$, all lower--orthant indicator monomials
$\phi_{i,s}(x)$ indexed by an observation $i\in[n]$ and a nonempty coordinate
subset $s\subseteq[d]$. We collect these features into blocks according to the
interaction order (equivalently, the number of coordinates involved), namely
$k:=|s|$. Writing the feature vector as a concatenation of blocks, we have
\[
h(x)
=
\bigl(h^{(1)}(x),\,h^{(2)}(x),\,\ldots,\,h^{(d)}(x)\bigr),
\qquad
h^{(k)}(x)
:=
\bigl(\phi_{i,s}(x): i\in[n],\ s\subseteq[d],\ |s|=k\bigr).
\]
In other words, $h^{(k)}(x)$ consists of all features that depend on exactly
$k$ coordinates of $x$ (all $k$-way interactions). Since there are $\binom{d}{k}$
choices of $s$ with $|s|=k$ and, for each such $s$, there are $n$ knots indexed
by $i\in[n]$, the block $h^{(k)}(x)$ has $n\binom{d}{k}$ entries. Therefore the
total number of \emph{non-intercept} features is
\[
p
=
\sum_{k=1}^d n\binom{d}{k}
=
n(2^d-1).
\]
An intercept can be included separately as an additional constant
feature.

The empirical design matrix $H\in\{0,1\}^{n\times p}$ is obtained by evaluating
the dictionary at each training point:
the $i$th row is $h(x_i)^\top$, $i=1,\ldots,n$.
We define the associated Gram (kernel) matrix
\[
K:=HH^\top\in\mathbb R^{n\times n},
\qquad
K_{ij}=h(x_i)^\top h(x_j)=:K(x_i,x_j).
\]
Since $K$ is symmetric and positive semidefinite, it admits an eigendecomposition $K=UDU^\top$, with $U$ orthogonal and $D=\mathrm{diag}(d_1,\ldots,d_n)$
satisfying $d_1\ge\cdots\ge d_n\ge0$.

A direct expansion gives, for generic $x,x'\in[0,\tau]^d$,
\[
\begin{aligned}
K(x,x')
&=
\sum_{i=1}^n\ \sum_{\emptyset\neq s\subseteq[d]}
\mathbf 1\{\,x_s\ge \tilde x_{i,s}\,\}\,\mathbf 1\{\,x'_s\ge \tilde x_{i,s}\,\} \\
&=
\sum_{i=1}^n\ \sum_{\emptyset\neq s\subseteq[d]}
\mathbf 1\{\, (x\wedge x')_s \ge \tilde x_{i,s}\,\},
\end{aligned}
\]
where $x\wedge x'$ denotes the componentwise minimum.
For each $i$, define the set of \emph{active coordinates} at the pair of points $(x,x')$
\[
S_i(x,x')
:=
\bigl\{\,j\in[d]:\ (x\wedge x')_j \ge \tilde x_{i,\{j\}}\,\bigr\}
=
\bigl\{\,j\in[d]:\ (x\wedge x')_j \ge x_{i,j}\,\bigr\}.
\]
Then the inner sum counts the nonempty subsets of $S_i(x,x')$, yielding
\[
K(x,x')
=
\sum_{i=1}^n \bigl(2^{|S_i(x,x')|}-1\bigr)
\]
which is the closed-form representation of the entries used in \citep{schuler2024highlyadaptiveridge}
and will serve as a computational building block below.

\section{PCHAL and PCHAR}
The introduction of the kernel $K$ in \cite{schuler2024highlyadaptiveridge}
is primarily computational. In our setting, however, its role is more
structural. The central question is how to efficiently organize and compress
the enormous collection of HAL basis functions. Our answer is to exploit
principal-component ideas, which motivates the names PCHAL and PCHAR.

Let $h(x)\in\mathbb R^p$ denote the vector of HAL basis functions evaluated at
$x$, and let
\[
H=
\begin{pmatrix}
h(X_1)^\top\\
\vdots\\
h(X_n)^\top
\end{pmatrix}
\in\mathbb R^{n\times p}
\]
be the corresponding design matrix on the training sample. Write a singular
value decomposition of $H$ as
\[
H = U D^{1/2} V^\top,
\]
where $U\in\mathbb R^{n\times n}$ and $V\in\mathbb R^{p\times n}$ have
orthonormal columns, and
$D=\mathrm{diag}(d_1,\ldots,d_n)$ with $d_1\ge \cdots \ge d_n\ge 0$.
For completeness, we also consider a truncated representation. For $k\le n$,
let $U_k$ and $V_k$ denote the first $k$ columns of $U$ and $V$, and let
$D_k$ be the leading $k\times k$ principal block of $D$.

We define the $k$-dimensional PC feature map by
\[
z_k(x):=V_k^\top h(x)\in\mathbb R^k.
\]
Evaluated on the training sample, this yields the PC score matrix
\[
Z_k:=
\begin{pmatrix}
z_k(X_1)^\top\\
\vdots\\
z_k(X_n)^\top
\end{pmatrix}
=HV_k
=U_kD_k^{1/2}
\in\mathbb R^{n\times k}.
\]
Since $K=HH^\top=UDU^\top$, the matrices $U$ and $D$ are exactly those
appearing in the eigendecomposition of $K$. Consequently, \(Z_k\) may be computed directly from \(K\), without explicitly forming \(H\) or \(V\), while still encoding the basis compression induced by \(HV_k\).

For a fixed truncation level \(k\) and regularization parameter \(\lambda \ge 0\), training in PC space amounts to regressing \(Y\) on \(Z_k\) by solving a penalized least-squares problem. In particular, PCHAR uses an \(\ell_2\) penalty of the form \(\lambda\|\beta\|_2^2\), whereas PCHAL uses an \(\ell_1\) penalty of the form \(\lambda\|\beta\|_1\). Thus, after projecting onto the leading \(k\) principal components, estimation reduces to a well-conditioned \(k\)-dimensional optimization problem while still preserving the geometry encoded by \(K\). Throughout, we use empirical least-squares loss for simplicity.

\begin{theorem}[PCHAL/PCHAR]\label{thm:PCHA}
For \(\lambda>0\):
\begin{itemize}
\item The ridge estimator
\[
    \hat\beta^{\mathrm{PCHAR}}_{k,\lambda}
      := \arg\min_{\beta\in\mathbb{R}^{k}}
         \Bigl\{\tfrac{1}{2n}\|Y-Z_k\beta\|_2^{2}
               +\tfrac{\lambda}{2}\|\beta\|_2^{2}\Bigr\}
\]
admits the closed form
\[
    \hat\beta^{\mathrm{PCHAR}}_{k,\lambda}
      \;=\; (D_{k}+n\lambda I_k)^{-1}\,D_{k}^{1/2}\,U_{k}^{\top}Y.
\]

\item The lasso estimator
\[
\hat\beta^{\mathrm{PCHAL}}_{k,\lambda}
:= \arg\min_{\beta\in\mathbb{R}^{k}}
\Bigl\{\tfrac{1}{2n}\|Y-Z_k\beta\|_2^{2} + \lambda\|\beta\|_1\Bigr\}
\]
admits the closed form
\[
\hat\beta^{\mathrm{PCHAL}}_{k,\lambda}
\;=\; D_{k}^{-1}\,
\mathrm{sign}(Z_k^{\top}Y)\,
\Bigl(\,|Z_k^{\top}Y|-n\lambda\Bigr)_+ ,
\]
for coordinates with \(d_j > 0\); coordinates with \(d_j=0\) are set to zero.
\end{itemize}
\end{theorem}

An alternative expression of the PCHAL coefficient, obtained by reparametrizing in the
orthonormal basis \(U_k\), is
\[
\hat \beta^{\mathrm{PCHAL}}_{k,\lambda}
\;=\;
D^{-1/2}_{k}\,
\mathrm{sign}(U_{k}^\top Y)\,
\bigl(|U_{k}^\top Y| - \lambda n D^{-1/2}_{k}\mathbf{1}_k\bigr)_+,
\]
where \(\mathbf{1}_k\) is the \(k\)-vector of ones and the absolute value and
\((\cdot)_+\) act componentwise. In particular, for each \(j\in\{1,\ldots,k\}\),
\[
\hat \beta^{\mathrm{PCHAL}}_{k,\lambda}(j) \neq 0
\quad\Longleftrightarrow\quad
\lambda < \frac{\sqrt{d_j}}{n}\,\bigl|u_j^\top Y\bigr|,
\]
where \(u_j\) is the \(j\)-th eigenvector of \(K\). Thus a cross–validation
procedure indexed by \(\lambda\) automatically selects the effective rank:
defining
\[
W_j := \frac{\sqrt{d_j}}{n}\,\bigl|u_j^\top Y\bigr|,
\]
the active principal components for a given \(\lambda\) are exactly
\[
\{\, j : W_j > \lambda \,\}.
\]
Sorting the values \(W_j\) in decreasing order yields a nested sequence of
models in which the number of selected components is monotonically decreasing
in \(\lambda\), in contrast to the standard lasso with correlated features.\\

The resulting fitted function is
\[
\widehat f_{k,\lambda}(x):=z_k(x)^\top \widehat\beta_{k,\lambda}.
\]
In particular, on the training sample,
\[
\bigl(\widehat f_{k,\lambda}(X_1),\ldots,\widehat f_{k,\lambda}(X_n)\bigr)^\top
= Z_k \widehat\beta_{k,\lambda}.
\]

\subsection{Tuning the number of principal components and the penalty Parameter}
For theoretical clarity, we present the estimators for fixed \((k,\lambda)\), although
in practice both hyperparameters are chosen adaptively. For each candidate \(k\),
let \(Z_k\) denote the full-sample \(k\)-dimensional score matrix, and for each fold
\(v=1,\dots,V\), let \(Z_k^{(-v)}\) and \(Z_k^{(v)}\) be the training- and validation-fold
submatrices obtained by restricting the rows of \(Z_k\) to the corresponding indices.
For each \(\lambda\) in a candidate grid \(\Lambda\), we fit
\(\widehat\beta_{k,\lambda}^{(-v)}\) on \((Z_k^{(-v)},Y^{(-v)})\) and evaluate
\[
\mathrm{CVRisk}(k,\lambda)
:=
\frac1V\sum_{v=1}^V \frac{1}{|I_v|}
\bigl\|Y^{(v)}-Z_k^{(v)}\widehat\beta_{k,\lambda}^{(-v)}\bigr\|_2^2.
\]
Thus, cross-validation is used only to tune \(\lambda\) \emph{within the fixed
\(k\)-dimensional basis}. For each \(k\), we define
\[
\hat\lambda(k)\in\arg\min_{\lambda\in\Lambda}\mathrm{CVRisk}(k,\lambda).
\]
We then refit the estimator on the full training sample at \((k,\hat\lambda(k))\), and
compare the resulting fits across \(k\) using the empirical training risk
\[
\widehat R\bigl(k,\hat\lambda(k)\bigr)
:=
\frac{1}{n}\Bigl\|Y-Z_k\widehat\beta_{k,\hat\lambda(k)}\Bigr\|_2^2.
\]
Accordingly, we select
\[
\hat k\in\arg\min_k \widehat R\bigl(k,\hat\lambda(k)\bigr),
\qquad
\hat\lambda:=\hat\lambda(\hat k).
\]
Finally, we refit on the full sample at \((\hat k,\hat\lambda)\) and define
\[
\widehat f^{\mathrm{PCHAL}}_{\hat k,\hat\lambda}(x)
:= z_{\hat k}(x)^\top \widehat\beta^{\mathrm{PCHAL}}_{\hat k,\hat\lambda},
\qquad
\widehat f^{\mathrm{PCHAR}}_{\hat k,\hat\lambda}(x)
:= z_{\hat k}(x)^\top \widehat\beta^{\mathrm{PCHAR}}_{\hat k,\hat\lambda}.
\]
\noindent

\paragraph{Remark.}
Our tuning rule is not fully honest, since the principal-component score basis is
constructed once from the full training sample and then reused throughout the
cross-validation step for selecting \(\lambda\). As a result, the validation-fold
predictions are evaluated in a representation that has already been informed by the
held-out covariates, so a mild form of data leakage is present at the feature-construction
stage. A fully honest alternative would recompute the principal-component basis inside
each training fold before evaluating validation risk. In our setting, however, this
foldwise recomputation is substantially more expensive when the covariate
dimension is large, while preliminary comparisons indicated only minor differences in predictive performance. For this reason, we adopt the
present tuning scheme in practice.

\subsection{Early-stopped gradient descent as soft PC regularization}
We propose an alternative PCHAL-type procedure based on implicit regularization through early-stopped gradient descent. Instead of selecting the number of retained principal components by cross-validation over a discrete grid of ranks, we use all PCs and cross-validate the stopping time. This replaces hard PC truncation by a smooth spectral regularization path. We expect this approach to be attractive when n is large, since tuning over stopping times can be computationally simpler than repeatedly fitting over many candidate PC ranks. Empirically, the early-stopped procedure and the PCHAL estimator achieve broadly similar predictive performance.\\

Let
\[
J = I_n - \frac{1}{n}\mathbf 1\mathbf 1^\top
\]
be the centering matrix. We work with the centered response
\[
Y_c = JY = Y-\bar Y\mathbf 1
\]
and the centered HAL kernel matrix
\[
K_c = J K J.
\]
Let
\[
K_c = UDU^\top
\]
be its eigendecomposition, where
\[
D=\operatorname{diag}(d_1,\ldots,d_r)
\]
contains the positive eigenvalues.
Instead of explicitly selecting a hard number of principal components, we use
all available PCs and regularize by early-stopped gradient descent. Starting
from
\[
\beta^{(0)}=0,
\]
we iterate
\[
\beta^{(t+1)}
=
\beta^{(t)}
-
\eta\frac{1}{n}Z^\top(Z\beta^{(t)}-Y_c),
\]
where \(\eta>0\) is the step size.

Because the PC score matrix is orthogonal in sample, the update separates
coordinatewise. For the \(j\)-th PC coordinate,
\[
\beta_j^{(t+1)}
=
\left(1-\eta\frac{d_j}{n}\right)\beta_j^{(t)}
+
\eta\frac{z_j^\top Y_c}{n},
\]
where \(z_j\) denotes the \(j\)-th column of \(Z\). The full least-squares
coefficient in this coordinate is
\[
\beta_j^{\mathrm{LS}}
=
\frac{z_j^\top Y_c}{d_j}.
\]
Thus gradient descent moves each coefficient geometrically from zero toward
its least-squares value:
\[
\beta_j^{(t)}
=
\left\{
1-\left(1-\eta d_j/n\right)^t
\right\}
\beta_j^{\mathrm{LS}}
=
\left\{
1-\left(1-\eta d_j/n\right)^t
\right\}
\frac{z_j^\top Y_c}{d_j}.
\]

Equivalently, the fitted values after \(t\) iterations are
\[
\widehat f_t
=
Z\beta^{(t)}
=
U\,\operatorname{diag}\{g_t(d_j)\}\,U^\top Y_c,
\]
where
\[
g_t(d_j)
=
1-\left(1-\eta d_j/n\right)^t
\]
is the spectral filter induced by \(t\) steps of gradient descent.

This formula shows how early stopping regularizes the PC-HAL fit. Directions
with large eigenvalues have large \(d_j/n\), so
\[
\left(1-\eta d_j/n\right)^t
\]
decays quickly and \(g_t(d_j)\) approaches one after only a few iterations.
Hence large-eigenvalue PC directions enter the fit rapidly. By contrast,
directions with small eigenvalues have small \(d_j/n\), so \(g_t(d_j)\)
increases slowly and these components enter the fit only after many
iterations. Early stopping therefore delays the contribution of
small-eigenvalue directions, which are the directions most susceptible to
noise amplification.

For interpretation, define the effective number of principal components used
at time \(t\) by
\[
k_{\mathrm{eff}}(t)
=
\sum_{j=1}^r g_t(d_j)
=
\sum_{j=1}^r
\left\{
1-\left(1-\eta d_j/n\right)^t
\right\}.
\]
This quantity counts each PC fractionally according to how much of its
least-squares coefficient has been learned by time \(t\). If
\(g_t(d_j)\approx 1\), the \(j\)-th PC is essentially included; if
\(g_t(d_j)\approx 0\), it is essentially excluded.

Thus early-stopped gradient descent replaces the explicit and discontinuous
choice of a rank \(k\) by a smooth optimization-time parameter \(t\). Small
\(t\) gives a heavily regularized low-complexity fit, while large \(t\)
approaches the full least-squares fit in the PC-HAL space.

In practice, the stopping time is selected by cross-validation. For a finite
grid \(\mathcal T\) of candidate stopping times, we define
\[
\widehat t
=
\arg\min_{t\in\mathcal T}
\widehat R_{\mathrm{CV}}(t),
\]
and the final centered fitted value is
\[
\widehat f_{\widehat t}
=
U\,\operatorname{diag}\{g_{\widehat t}(d_j)\}\,U^\top Y_c.
\]
The final prediction adds back the empirical mean of the response:
\[
\widehat Y
=
\bar Y\mathbf 1+\widehat f_{\widehat t}.
\]
Therefore the stopping time \(t\) plays the role of a regularization
parameter: we stop when validation risk is minimized, not when gradient
descent has converged.

\subsection{Prediction}
Prediction in the PCHAL framework requires some care.
Although the fitted predictor is linear in the retained PC scores, those scores are
defined through the singular value decomposition of the high-dimensional HAL design
matrix \(H\in\mathbb{R}^{n\times p}\), where \(p=n(2^d-1)\). Given new covariate points
\(X'_1,\ldots,X'_N\), the most direct approach would be to first construct their full
HAL feature matrix
\[
H'=
\begin{pmatrix}
h(X'_1)^\top\\
\vdots\\
h(X'_N)^\top
\end{pmatrix}
\in\mathbb{R}^{N\times p},
\]
and then project these expanded features onto the training PC directions, yielding
\[
Z'_k = H'V_k,
\qquad\text{and hence}\qquad
\hat f' = Z'_k\hat\beta = H'V_k\hat\beta,
\]
where \(V_k\in\mathbb{R}^{p\times k}\) contains the leading \(k\) right singular vectors
of the training design matrix and \(\hat\beta\in\mathbb{R}^k\) denotes the estimated
coefficient vector in PC space. This construction is algebraically natural, since it
simply applies to the new sample the same projection used to define the training PC
scores. However, it is computationally infeasible in the HAL setting, because the
ambient feature dimension \(p\) is extremely large, so neither the expanded matrix
\(H'\) nor the loading matrix \(V_k\) can realistically be formed or stored explicitly.
For this reason, prediction must instead be expressed through the corresponding
training--test Gram matrix, which yields the same projected scores without ever
constructing the full HAL feature expansion.

To obtain a computable prediction map, note that the singular value decomposition
of the training design implies
\[
V = H^\top U D^{-1/2}.
\]
Restricting to the leading \(k\) components gives
\[
H'V_k
=
H'H^\top U_k D_k^{-1/2}.
\]
Now define the \emph{cross-kernel} matrix
\[
K' := H'H^\top \in \mathbb{R}^{N\times n},
\]
whose \((i,j)\) entry is
\[
K'_{ij} = h(X'_i)^\top h(X_j).
\]
It follows that
\[
H'V_k = K'U_kD_k^{-1/2},
\qquad\text{and therefore}\qquad
f' = (K'U_kD_k^{-1/2})\hat\beta.
\]
Thus, prediction reduces to computing the training--test kernel matrix \(K'\), which
admits the same closed-form evaluation as the training Gram matrix \(K=HH^\top\).
This avoids any explicit construction of the high-dimensional objects \(H'\) and \(V_k\). In practice, we do not penalize the intercept. Accounting
for this choice introduces additional centering operations and makes the notation more
cumbersome, so we defer the precise formulation in terms of the centered design and
kernel matrices (that is, \(\tilde H\) and \(\tilde K\)) to Appendix~\ref{sec:center-ha-design}.

\subsection{Tuning the Interaction Order}
As the closed-form expression for \(K\) shows, rather than explicitly enumerating the large collection of sectional indicator basis functions as covariates, the effect of dimensionality is absorbed into the entries of the Gram matrix \(K\). When the number of covariates becomes very large, it is desirable to control the maximum interaction order in order to avoid numerical instability and reduce computational burden, although in most of the practical settings this is not a serious issue. \\

This observation also suggests a natural extension of the kernel construction: if we restrict the dictionary to interactions of order at most \(m\),
\[
h_{\le m}(x):=\bigl(h^{(1)}(x),\ldots,h^{(m)}(x)\bigr),
\qquad
K_{\le m}(x,x'):=h_{\le m}(x)^\top h_{\le m}(x'),
\]
then for each $i$ the number of retained subsets is $\sum_{\ell=1}^m\binom{|S_i(x,x')|}{\ell}$, so
\[
K_{\le m}(x,x')
=
\sum_{i=1}^n \ \sum_{\ell=1}^m \binom{|S_i(x,x')|}{\ell},
\]
as derived in Appendix \ref{sec:truncha}.\\

In this subsection, we fix the number of principal components and consider a collection of estimators indexed by two hyperparameters: a regularization parameter \(\lambda \in \Lambda\) where $|\Lambda|<\infty$ and a maximum interaction degree \(m \in \{0,1,\ldots,M\}\). For each pair \((m,\lambda)\), let \(\hat{f}_{m,\lambda}\) denote the PCHA estimator obtained by using all basis elements with interaction degree at most \(m\), together with penalty level \(\lambda\), while keeping the rank fixed. The resulting candidate library is the set
\[
\mathcal{F}
\;=\;
\bigl\{\hat{f}_{m,\lambda} : m \in \{0,\ldots,M\},\; \lambda \in \Lambda\bigr\},
\]
and we denote its cardinality by \(I := |\mathcal{F}|\). Throughout this subsection, we assume that both \(M\) and the grid \(\Lambda\) are fixed and do not depend on \(n\), so that \(I\) remains fixed as \(n \to \infty\).

Under standard regularity conditions for \(V\)-fold cross-validation, the oracle results of
\citet{dudoit2005asymptotics} and \citet{vdlaan_dudoit_vandervaart} imply that the
cross-validated selector is asymptotically oracle-optimal in risk over the finite library
\(\mathcal F\). We now state this more carefully.

\medskip

For each pair \((m,\lambda)\), let \(\hat f^{(n)}_{m,\lambda}\) denote the estimator trained on a sample
of size \(n\). Define its prediction risk under squared error loss by
\[
R_n(m,\lambda)
\;:=\;
\mathbb E\!\left[
\bigl(Y_{n+1}-\hat f^{(n)}_{m,\lambda}(X_{n+1})\bigr)^2
\right],
\]
where the expectation is taken over both the training sample and an independent test point
\((X_{n+1},Y_{n+1}) \sim P_0\). The oracle selector is any minimizer
\[
(m_n^\ast,\lambda_n^\ast)\in \arg\min_{m,\lambda} R_n(m,\lambda),
\]
which depends on the unknown data-generating distribution \(P_0\).

Cross-validation provides an empirical estimator \(\widehat R_{\mathrm{CV}}(m,\lambda)\) of
\(R_n(m,\lambda)\) by fitting \(\hat f_{m,\lambda}\) on training folds and evaluating squared
prediction error on validation folds. The resulting cross-validated selector is
\[
(\widehat m,\widehat\lambda)
\;=\;
\arg\min_{m,\lambda}\,\widehat R_{\mathrm{CV}}(m,\lambda).
\]

The oracle inequalities in the above references imply, in particular, that
\[
R_n(\widehat m,\widehat\lambda)
-
\inf_{m,\lambda} R_n(m,\lambda)
\;=\;
o_p(1).
\]
Moreover, for a finite candidate library, the complexity term in the corresponding oracle bound
depends at most logarithmically on the library size. Since in our setting
\(I=|\mathcal F|\) is fixed, this term is asymptotically negligible. Thus the
cross-validated selector performs asymptotically as well, in risk, as the oracle selector over
\(\mathcal F\).

For each interaction degree \(m\), define the \emph{profiled} cross-validated risk
\[
\widehat{R}_{\mathrm{CV}}(m)
\;:=\;
\min_{\lambda \in \Lambda}\widehat{R}_{\mathrm{CV}}(m,\lambda),
\]
where the minimization is taken over the fixed grid of regularization parameters.
If one wishes to perform a global cross-validated selection over the full candidate
library, then one would choose
\[
(\widehat{m}_{\mathrm{glob}},\widehat{\lambda}_{\mathrm{glob}})
\;=\;
\arg\min_{(m,\lambda)\in \{0,\ldots,M\}\times\Lambda}
\widehat{R}_{\mathrm{CV}}(m,\lambda),
\]
which is equivalently obtained by first profiling over \(\lambda\) and then minimizing over \(m\):
\[
\widehat{m}_{\mathrm{glob}}
\;=\;
\arg\min_{m\in\{0,\ldots,M\}} \widehat{R}_{\mathrm{CV}}(m),
\qquad
\widehat{\lambda}_{\mathrm{glob}}
\in
\arg\min_{\lambda\in\Lambda}\widehat{R}_{\mathrm{CV}}(\widehat{m}_{\mathrm{glob}},\lambda).
\]
Thus, profiling over \(\lambda\) is exactly equivalent to joint minimization over the
full library.

In practice, however, one may instead adopt a simpler \emph{forward-complexity}
selection rule for \(m\). Starting from a small interaction degree, we compare the
profiled risks sequentially and stop at the first value of \(m\) for which increasing
the interaction degree no longer decreases the profiled CV risk. That is, we define
\[
\widehat{m}_{\mathrm{for}}
\;:=\;
\min\Bigl\{m\in\{0,\ldots,M-1\}:
\widehat{R}_{\mathrm{CV}}(m+1)\ge \widehat{R}_{\mathrm{CV}}(m)\Bigr\},
\]
with the convention that \(\widehat{m}_{\mathrm{for}}=M\) if no such \(m\) exists.
This forward rule is generally \emph{not} identical to the global selector
\(\widehat{m}_{\mathrm{glob}}\), since it may stop at the first local failure of the
profiled risk to decrease, even though a larger interaction degree could in principle
achieve a smaller CV risk later. Thus, the forward rule should be viewed as a modified,
computationally convenient selector of interaction complexity, rather than as the exact
global minimizer over all \(m\).

The corresponding oracle version of this forward rule is obtained by replacing the
profiled CV risk with the profiled population risk
\[
R_n^\dagger(m)
\;:=\;
\min_{\lambda\in\Lambda} R_n(m,\lambda),
\]
where \(R_n(m,\lambda)\) denotes the prediction risk of the estimator trained on a
sample of size \(n\). The forward oracle selector is then
\[
m_{\mathrm{for}}^*(n)
\;:=\;
\min\Bigl\{m\in\{0,\ldots,M-1\}: R_n^\dagger(m+1)\ge R_n^\dagger(m)\Bigr\},
\]
again with the convention \(m_{\mathrm{for}}^*(n)=M\) if no such \(m\) exists.
If the true regression function is not contained in any model with finite interaction
degree \(m\le M\), then this oracle index will typically depend on \(n\): as the sample
size increases, richer interaction structures may become worthwhile because their
approximation error decreases while the variance cost becomes more manageable.
Accordingly, the target interaction degree is itself sample-size dependent.

For each fixed \(m\), the inner minimization over \(\lambda\) is still performed over
the full grid \(\Lambda\), so that the profiled procedure at that complexity level is
oracle-optimal relative to \(\lambda\). The forward selector then compares these
profiled risks across successive values of \(m\). It is therefore natural to view the
cross-validated forward rule as aiming to track the forward oracle
\(m_{\mathrm{for}}^*(n)\). A proof that this sequential selector is asymptotically
equivalent to the \emph{global} oracle selector would require additional assumptions on
the shape of the profiled risk curve \(m \mapsto R_n^\dagger(m)\), for example that
once the risk ceases to decrease it cannot later decrease again. We do not pursue such
conditions here.

A small simulation study illustrating how cross-validation selects the interaction order is reported in Appendix~\ref{appendix:interaction-order-cv}. In particular, the experiment compares the cross-validated selector with the corresponding oracle selector and shows how the selected interaction degree evolves with sample size.

\subsection{Explicit form of the PC scores}\label{sec:d1}
Since the entries of \(H\) are essentially binary, the matrix \(H\) already possesses a highly structured form. This motivates a closer examination of the associated Gram matrix \(K\). Interestingly, in certain special cases, \(K\) admits an exact identification as the covariance matrix of a random walk. \\
When $d=1$, the construction becomes deterministic once the sample is sorted.
Let $x_{(1)}\le\cdots\le x_{(n)}$ denote the order statistics. Constructing the
HAL design matrix in this ordered basis yields
\[
H_{ij}=\mathbf 1\{x_{(i)}\ge x_{(j)}\}=\mathbf 1\{i\ge j\},
\qquad 1\le i,j\le n,
\]
so $H$ is a unit lower-triangular matrix of ones. Its Gram matrix satisfies
\[
(HH^\top)_{ij}=\sum_{k=1}^{n} H_{ik}H_{jk}
=\sum_{k=1}^{\min(i,j)}1=\min(i,j),
\qquad 1\le i,j\le n.
\]
Consequently,
\[
\begin{minipage}{0.45\textwidth}\centering
$
H=
\begin{pmatrix}
1&0&0&\cdots&0\\
1&1&0&\cdots&0\\
1&1&1&\cdots&0\\
\vdots&\vdots&\vdots&\ddots&\vdots\\
1&1&1&\cdots&1
\end{pmatrix}
$
\end{minipage}
\qquad
\begin{minipage}{0.45\textwidth}\centering
$
HH^\top=
\begin{pmatrix}
1&1&1&\cdots&1\\
1&2&2&\cdots&2\\
1&2&3&\cdots&3\\
\vdots&\vdots&\vdots&\ddots&\vdots\\
1&2&3&\cdots&n
\end{pmatrix}
$.
\end{minipage}
\]

In dimensions $d>1$, an explicit description of the eigenvectors of the zero-order
HAL Gram matrix is generally unavailable. The following theorem identifies a
tractable special case: when the sample can be totally ordered under the
coordinatewise partial order.

\begin{theorem}[Eigenstructure of the zero-order HAL Gram matrix]\label{thm:hal_gram_eigs}
Assume there exists a permutation $\pi$ of $\{1,\ldots,n\}$ such that, for the
reordered sample $X_{(i)}:=X_{\pi(i)}$, each coordinate is strictly increasing in $i$:
\[
X_{(1),j} < X_{(2),j} < \cdots < X_{(n),j},
\qquad j=1,\ldots,d.
\]
Let $H$ be the zero-order indicator HAL design matrix built from all lower-orthant
monomials indexed by nonempty coordinate subsets (no intercept), evaluated at
$X_{(1)},\ldots,X_{(n)}$, and let $K:=HH^\top\in\mathbb{R}^{n\times n}$ be the
associated Gram matrix. Then
\[
K \;=\; (2^d-1)\,A,
\qquad
A_{ij}=\min(i,j),\ \ 1\le i,j\le n.
\]
Equivalently, in the original indexing,
\[
K \;=\; P^\top\!\bigl((2^d-1)A\bigr)P,
\]
where $P$ is the permutation matrix corresponding to $\pi$.

Consequently, the eigenvectors of $K$ are obtained by permuting those of $A$.
In particular, an orthonormal eigenbasis of $A$ is given by the discrete sine
vectors
\[
u_{k}(i)
=
\sqrt{\frac{4}{2n+1}}\,
\sin\!\left(\frac{(2k-1)\,i\,\pi}{2n+1}\right),
\qquad
k=1,\dots,n,\ \ i=1,\dots,n,
\]
with eigenvalues
\[
\lambda_k(A)
=
\frac{1}{2\Bigl(1-\cos\!\bigl(\frac{(2k-1)\pi}{2n+1}\bigr)\Bigr)}
=
\frac{1}{4\sin^2\!\bigl(\frac{(2k-1)\pi}{4n+2}\bigr)}.
\]
Therefore,
\[
\lambda_k(K)=(2^d-1)\lambda_k(A),
\qquad
\text{and}\qquad
\sigma_k(H)=\sqrt{\lambda_k(K)}
=
\sqrt{2^d-1}\,
\frac{1}{2\sin\!\Bigl(\frac{(2k-1)\pi}{4n+2}\Bigr)}.
\]
\end{theorem}
There is a striking connection between our Gram matrix and classical Gaussian-process theory. In the special case considered here, the relevant eigensystem is exactly that of the covariance matrix of a random walk, with continuum analogue given by the covariance kernel \(\min(s,t)\) of Brownian motion. The latter has the well-known Karhunen--Lo\`eve expansion in the sine basis, and the discrete matrix \(A_{ij}=\min(i,j)\) inherits the corresponding discrete sine eigenstructure \citep{pavliotis2014stochastic}. A proof of this discrete eigensystem, together with further discussion, is given in \citep{trench1999eigenvalues}; for completeness, we also provide a self-contained derivation in Appendix~\ref{app:hal-gram-eigs}. Figure~\ref{fig:eigenvectors} overlays the leading eigenvectors of \(HH^\top\) with their closed-form discrete sine counterparts.
\begin{figure}[H]
	\centering
	\includegraphics[width=0.75\textwidth]{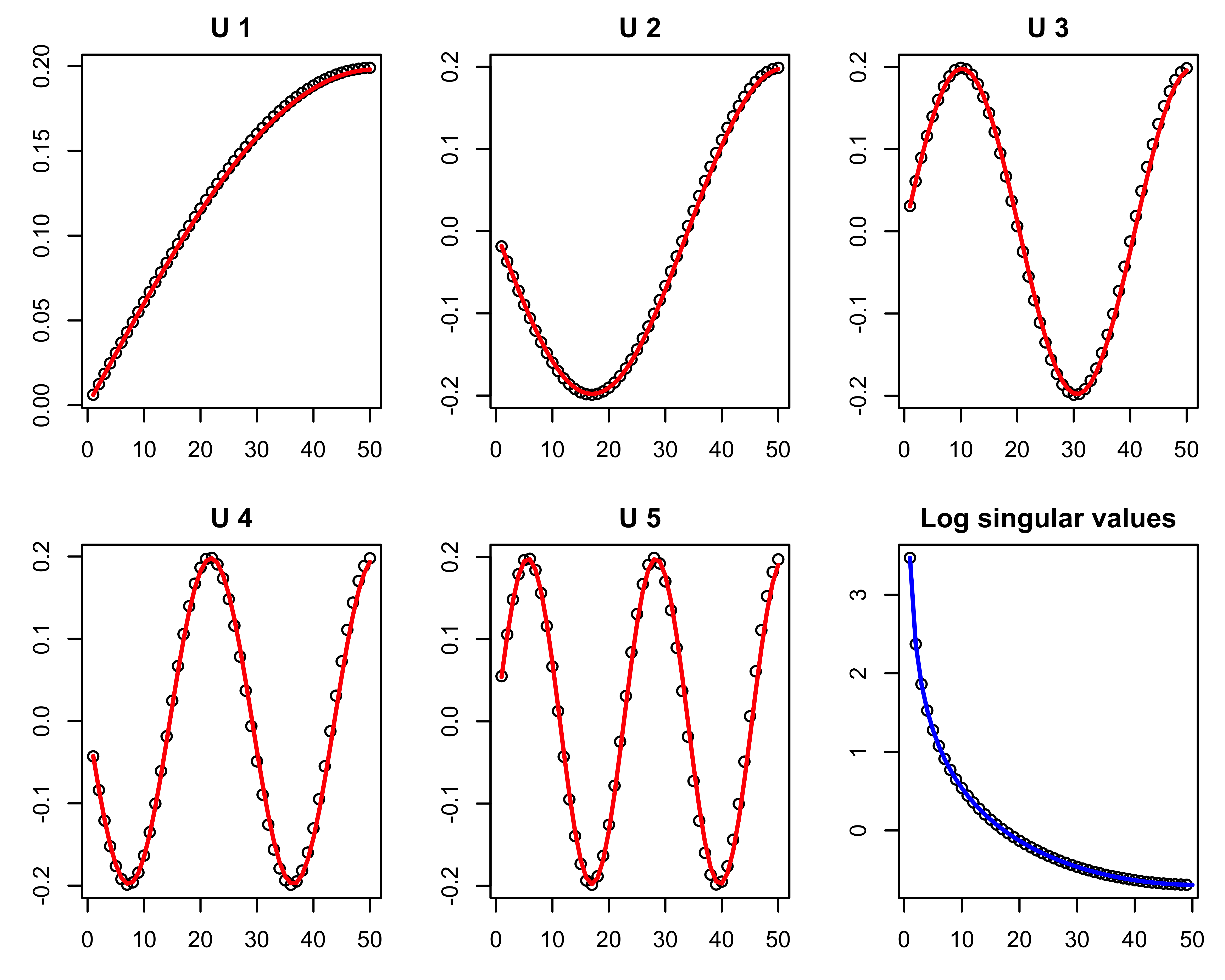}
	\caption{First six eigenvectors of $HH^\top$ (black) if there is a total order, overlaid with their closed-form discrete sine eigenfunctions (red).}
	\label{fig:eigenvectors}
\end{figure}

\section{Empirical Performance}
\subsection{Experimental setup}

We evaluate the proposed PCHA estimators against several popular learners on a collection of real-world regression datasets \citep{dua2017uci}. We refer to \citep{schuler2024highlyadaptiveridge} for reported performance of HAL on the same datasets; due to its computational burden, we do not include HAL directly here. For each
dataset, we retain up to $N_{\max}=2000$ complete observations (dropping rows
with missing values). Within each repetition, we split the data into
an $80\%/20\%$ train/test partition, apply train-fitted min--max scaling to map
each feature to $[0,1]$, and evaluate predictive accuracy by test RMSE. Results
are averaged over $R=5$ random splits. 
\subsection{HAL-kernel and PC--HA estimators}

All HAL-family estimators in this real-data experiment are implemented through
the zero-order highly adaptive regression kernel. Thus we do not explicitly
construct the full HAL basis matrix $H$. Instead, we compute the training Gram
matrix
\[
K = HH^\top
\]
and the test--training cross-kernel
\[
K' = H'H^\top
\]
using the closed-form formula proposed in \citep{schuler2024highlyadaptiveridge}. Before fitting, the kernel is normalized by
the mean diagonal of $K$, and the train and cross kernels are centered using
the training empirical centering operation. The response is centered during
kernel fitting, and the intercept is added back as the training-sample mean.

\begin{itemize}
  \item \textbf{HAR-kernel ridge.}
  HAR is fit as kernel ridge regression with the centered highly adaptive
  kernel. For each training fold and for the final fit, we solve
  \[
    (K_c+n\lambda I)\alpha = Y_c,
  \]
  where $K_c$ is the centered and normalized training kernel and $Y_c$ is the
  centered response. The regularization parameter is selected by 5-fold
  cross-validation over
  \[
    \lambda \in 10^{\{-8,-7,\ldots,2\}}.
  \]

  \item \textbf{PCHAR.}
  Candidate dimensions are chosen on a 10-point grid from approximately
  $\max\{5,\lfloor n_{\rm tr}/10\rfloor\}$ to $n_{\rm tr}-1$, truncated by the
  numerical rank. For each $k$, $\lambda$ is selected by 5-fold cross-validation
  on
  \[
    \lambda \in 10^{\{1,\ldots,-9\}},
  \]
  using 25 equally spaced log-scale values. Conditional on $(k,\lambda)$, the
  ridge coefficient is computed by the closed form
  \[
    \hat\beta_{k,\lambda}
    =
    \frac{Z_k^\top Y_c}{D_k+n\lambda}.
  \]
  The dimension $k$ is selected by the resulting training MSE, and then
  $\lambda$ is reselected by 5-fold cross-validation at the chosen $k$.

  \item \textbf{PCHAL.}
  PCHAL uses the same centered kernel, eigendecomposition, candidate $k$ grid,
  and $\lambda$ grid as PCHAR, but replaces the ridge penalty with an
  $\ell_1$ penalty in the PC coordinates. Since the columns of $Z_k$ are
  orthogonal, the lasso coefficients are computed by soft-thresholding:
  \[
    \hat\beta_{k,\lambda,j}
    =
    \frac{
    \operatorname{sign}(Z_{k,j}^\top Y_c)
    \left(|Z_{k,j}^\top Y_c|-n\lambda\right)_+
    }{d_j},
    \qquad j=1,\ldots,k.
  \]
  As with PCHAR, $k$ is selected by training MSE after cross-validating
  $\lambda$ for each candidate $k$, and $\lambda$ is then reselected at the
  final chosen dimension.
\end{itemize}

\subsection{Baseline learners}
We compare the HAL-kernel methods with the following baseline learners.

\begin{itemize}
  \item \textbf{Ordinary ridge regression.}
  Ridge regression is fit directly on the scaled covariates with an intercept.
  The penalty parameter is selected by 5-fold cross-validation over a dense
  logarithmic grid of regularization values. No additional rescaling is applied
  beyond the train-fitted min--max normalization described above.

  \item \textbf{Gaussian kernel ridge regression.}
  We use the radial basis function (RBF) kernel
  \[
    K(x,x')=\exp\{-\sigma\|x-x'\|^2\}.
  \]
  The ridge penalty is selected from
  \[
    \lambda \in 10^{\{-6,-5,-4,-3,-2,-1\}},
  \]
  while the kernel bandwidth is chosen from
  \[
    \sigma \in \{0.25,0.5,1,2,4\}/\widehat m,
  \]
  where $\widehat m$ denotes the median squared pairwise distance among the
  training observations. Thus, $\sigma$ controls the locality of the kernel,
  with larger values yielding more localized fits. The pair $(\lambda,\sigma)$
  is selected by 5-fold cross-validation.

  \item \textbf{$k$-nearest neighbors.}
  The number of neighbors is selected by 5-fold cross-validation over
  \[
    k \in \mathrm{unique}\!\left(
    \mathrm{round}\{\sqrt n/2,\sqrt n,2\sqrt n,n^{2/3}/3,n^{2/3}\}
    \right),
  \]
  truncated to satisfy $2\le k<n$. The parameter $k$ controls the amount of
  smoothing: smaller values produce more local and variable fits, whereas larger
  values yield smoother but potentially more biased predictions.

  \item \textbf{Random forest.}
  Random forests are fit as ensembles of 800 regression trees. At each split,
  only a random subset of covariates is considered, with subset size equal to
  $1$ in the univariate case and $\lceil \sqrt p \rceil$ otherwise. Each tree is
  grown on a bootstrap-style subsample containing approximately $63.2\%$ of the
  training observations, and terminal node size is lower bounded by
  \[
    \max\{5,\lfloor n/200\rfloor\}.
  \]
  These choices control tree depth and local adaptivity while stabilizing the
  ensemble through averaging.

  \item \textbf{Generalized additive models (GAM).}
  GAMs are fit using additive spline smoothers. For covariates with at most five
  distinct observed values, we use a linear effect; otherwise, we fit a
  thin-plate spline smoother $s(X_j,k_j)$ with basis dimension
  \[
    k_j=\min\{60,\max(5,\lfloor n/10\rfloor),u_j-1\},
  \]
  where $u_j$ is the number of unique values of the $j$th covariate in the
  training sample. The smoothing parameters are estimated by restricted maximum
  likelihood (REML). This produces a flexible nonlinear additive model, while
  remaining additive across coordinates.

  \item \textbf{MARS.}
  Multivariate adaptive regression splines (MARS) are fit using piecewise linear
  hinge basis expansions. We cross-validate over the maximum interaction order,
  taking values
  \[
    1 \quad\text{or}\quad 2,
  \]
  over the pruning penalty
  \[
    \{1,1.2,1.5,2,3\},
  \]
  and over 10 candidate upper bounds on the number of basis terms, ranging from
  \[
    \max\{16,\lfloor 0.10n\rfloor\}
    \quad\text{to}\quad
    \min\{250,\lfloor 0.85n\rfloor\}.
  \]
  The final model is then refit on the full training sample using the selected
  tuning parameters.

  \item \textbf{XGBoost.}
  XGBoost is fit under squared-error loss using histogram-based tree
  construction. We tune the learning rate,
  \[
    \eta\in\{0.05,0.10,0.20\},
  \]
  the maximum tree depth,
  \[
    \{3,5,7\},
  \]
  and the minimum child weight,
  \[
    \{1,5\},
  \]
  by 5-fold cross-validation. Here, the learning rate controls the step size of
  boosting updates, the maximum depth controls tree complexity, and the minimum
  child weight regularizes splits by requiring sufficient information in each
  child node. We fix the row and column subsampling fractions at $0.8$, use an
  $\ell_2$ penalty of $1$, allow up to 1200 boosting rounds, and apply early
  stopping with patience 60. The final model is refit using the selected number
  of rounds and tuning configuration.
\end{itemize}

Table~\ref{tab:real} reports mean test RMSE for each dataset (best method per
row in bold). A clear pattern is that the PC-based HA estimators (PCHAR/PCHAL)
behave \emph{stably} across datasets: they are rarely the very best method, but
they also avoid severe degradations and typically remain close to the full
HA-kernel benchmark (e.g., the three HA variants are essentially
indistinguishable on \texttt{yearmsd}). In contrast, several competing learners
exhibit more pronounced dataset-to-dataset variability: they can achieve the
lowest RMSE on some problems (e.g., tree/boosting methods on \texttt{blog},
\texttt{energy}, \texttt{power}, \texttt{wine}, \texttt{yacht}, \texttt{yearmsd})
yet perform noticeably worse on others (e.g., RKRR is excellent on
\texttt{kin8nm} but substantially worse on \texttt{blog}/\texttt{yearmsd}).
Similarly, strongly structured models can be highly competitive when their
assumptions align with the data (e.g., GAM on \texttt{naval}) but less so
elsewhere. A small number of entries are omitted on very wide datasets due to
practical runtime constraints and are shown as missing in the table.

\begin{table}[H]
\centering

\scriptsize
\setlength{\tabcolsep}{3pt}
\renewcommand{\arraystretch}{1.05}
\resizebox{\textwidth}{!}{%
\begin{tabular}{lrrcccccccccc}
\hline
data & $n$ & $p$ & HAR & PCHAR & PCHAL & RKRR & RF & Ridge & kNN & GAM & MARS & XGB\\
\hline
blog    & 2000 & 266 & 7.72 & 7.73 & 8.50 & 1.23e+1 & 1.69e+1 & 2.08e+1 & --- & --- & --- & \textbf{6.82}\\
boston  & 506  & 13  & 3.53 & 3.53 & 3.68 & 3.20    & 3.33    & 4.85    & 5.48 & 3.67 & 3.74 & \textbf{3.10}\\
concrete& 1030 & 8   & \textbf{3.82} & 4.83 & 3.90 & 5.53 & 5.34 & 1.04e+1 & 9.85 & 5.39 & 5.25 & 4.06\\
energy  & 768  & 8   & 3.75e-1 & 3.97e-1 & 3.91e-1 & 6.26e-1 & 8.11e-1 & 2.87 & 2.59 & 1.04 & 4.86e-1 & \textbf{2.98e-1}\\
kin8nm  & 2000 & 8   & 1.41e-1 & 1.41e-1 & 1.50e-1 & \textbf{9.06e-2} & 1.73e-1 & 2.01e-1 & 1.47e-1 & 1.98e-1 & 1.76e-1 & 1.46e-1\\
naval   & 2000 & 15  & 8.98e-4 & 8.96e-4 & 1.05e-3 & 7.36e-4 & 1.21e-3 & 4.48e-3 & 6.67e-3 & \textbf{2.59e-5} & 4.55e-4 & 8.10e-4\\
power   & 2000 & 4   & 4.03 & 4.21 & 4.09 & 4.27 & 4.02 & 4.58 & 4.47 & 4.21 & 4.31 & \textbf{3.90}\\
protein & 2000 & 9   & 1.81 & 1.84 & 1.81 & 2.08 & \textbf{1.79} & 2.30 & 2.15 & 1.85 & 1.85 & 1.87\\
wine    & 1599 & 11  & 6.11e-1 & 6.11e-1 & 6.42e-1 & 6.42e-1 & 5.94e-1 & 6.60e-1 & 6.70e-1 & 6.51e-1 & 6.62e-1 & \textbf{5.91e-1}\\
yacht   & 308  & 6   & 7.28e-1 & 7.28e-1 & 7.35e-1 & 8.06e-1 & 2.08 & 8.34 & 9.19 & 1.69 & 1.15 & \textbf{5.38e-1}\\
yearmsd & 2000 & 90  & 1.16e+1 & 1.16e+1 & 1.16e+1 & 9.21e+1 & 1.01e+1 & 1.01e+1 & --- & --- & --- & \textbf{9.23}\\
\hline
\end{tabular}%
}
\caption{Real Dataset Mean RMSE}
\label{tab:real}
\end{table}

\section{Discussion}

Recent theoretical work on principal-component highly adaptive estimators
shows that PC-based versions of HAL can attain the same loss-based convergence
rate as HAL under suitable complexity-control assumptions \citep{garciameixide2026highlyadaptiveermpc}.
In particular, for a $k$th-order HAL basis with $k^*=k+1$, the corresponding
PC-HA estimator can achieve
\[
d_0(\hat\psi_{\mathrm{PC\text{-}HA}},\psi_0)
=
O_P^+\!\left(n^{-2k^*/(2k^*+1)}\right),
\]
and for zero-order HAL this gives the familiar loss-based rate
\[
O_P^+(n^{-2/3}),
\]
or equivalently an $L^2$-type rate $O_P^+(n^{-1/3})$ up to logarithmic factors.
Thus, under the appropriate assumptions, the principal-component reduction does
not necessarily sacrifice the statistical rate that motivates HAL in the first
place.

The main idea of PCHAL and PCHAR is to perform an outcome-blind compression of
the HAL basis. The principal components are determined only by the covariates,
through the HAL Gram matrix $K=HH^\top$, and do not use the response $Y$.
Equivalently, the construction depends only on the relative positions of the
observed covariates in the lower-orthant indicator geometry. This gives a
geometry-driven reorganization of the large HAL dictionary before regression is
performed. After this reduction, PCHAR becomes ridge regression in the PC score
space, while PCHAL becomes lasso regression in an orthogonal design, leading to
closed-form shrinkage rules.

Empirically, PCHAL and PCHAR behave similarly to
HAL and HAR while substantially reducing the computational burden of HAL \citep{schuler2024highlyadaptiveridge}. The early-stopped gradient descent
version is also convenient, since it avoids an explicit hard choice of the
number of principal components. Instead, the stopping time induces a smooth
spectral regularization path: large-eigenvalue directions enter the fit early,
while small-eigenvalue directions are delayed.

Finally, the spectral structure of the HAL Gram matrix reveals an interesting
connection with Brownian motion in special cases. When the observations are
totally ordered coordinatewise, the zero-order HAL Gram matrix reduces to a
scalar multiple of the matrix with entries
\[
A_{ij}=\min(i,j),
\]
which is the discrete covariance matrix of a random walk and the finite-sample
analogue of the Brownian motion covariance kernel $\min(s,t)$. In this setting,
the eigenvectors are discrete sine functions, matching the classical
Karhunen--Loève expansion of Brownian motion. Although this exact structure is
special, it gives a useful interpretation of PC-HAL: the method is extracting
low-frequency spectral modes from the geometry of the HAL basis.


\newpage
\appendix

\section{Proof of Theorem~\ref{thm:sectional_rep}}
\label{app:sectional-representation}
\par The proof is a slight modification of \citep{fang2021hk}.
\begin{definition} 
A function $f$ on $[0,1]^d$ is called \textbf{completely monotone} \citep{aistleitner2014functions} if for any closed axis-parallel box $A \subset [0,1]^d$ of arbitrary dimension $s$ (where $1 \leq s \leq d$), its $s$-dimensional quasi-volume generated by the function $f$ is non-negative. 
\end{definition}

\begin{lemma}
    For $V(f) < \infty$, $f(x) = f_1(x) - f_2(x)$ where $f_1(x)$ and $f_2(x)$ are completely monotone on $[0,1]^d$.
\end{lemma}
This is a multivariate extension of the Jordan decomposition on $\mathbb{R}$. While any function of bounded variation admits a decomposition into the difference of two completely monotone functions $f = f_1 - f_2$ (Leonov 1996 )\citep{leonov1996total}, a naive choice such as $f_1(x) = \text{Var}_{HK0}(f; [0,x])$ and $f_2(x) = f_1(x) - (f(x) - f(0))$ is generally insufficient for our purposes. 

Specifically, in $d$ dimensions, the function $f_2$ defined this way is not guaranteed to be completely monotone, meaning its quasi-volumes over sub-boxes may be negative. Furthermore, even if $f_1$ and $f_2$ were monotone, such a decomposition is generally not \textbf{unique} and does not satisfy the property that the total variation of the function equals the sum of the variations of its components. We are interested in the \textbf{canonical Jordan decomposition}, which is the unique representation $f = f(0) + f^+ - f^-$ anchored at zero that satisfies the variation identity:
\[
V(f) = V(f^+) + V(f^-).
\]
This specific decomposition is essential because it ensures a one-to-one correspondence with the Jordan decomposition of the signed measure $\mu_f$. The existence of this representation is established in the following theorem.

\begin{theorem}
    Let $f$ be a cadlag function on $[0,1]^d$ with bounded HK variation. Then there exist two uniquely determined completely monotone functions $f^+$ and $f^-$ on $[0,1]^d$ such that $f^+(0) = f^-(0) = 0$ and 
    \[
    f(x) = f(0) + f^+(x) - f^-(x), \quad x \in [0,1]^d,
    \]
    and
    \[
    V(f) = V(f^+) + V(f^-).
    \]
\end{theorem}

\begin{align*}
    f^+(x) &= \frac{1}{2} \left( \text{Var}_{\text{HK0}}(f;[0,x]) + f(x) - f(0) \right), \\
    f^-(x) &= \frac{1}{2} \left( \text{Var}_{\text{HK0}}(f;[0,x]) - f(x) + f(0) \right).
\end{align*}
Aistleitner and Dick \citep{aistleitner2014functions} proved that the two functions are completely monotone.

\begin{theorem}
    Let $f$ be a cadlag function on $[0,1]^d$ with $V(f) < \infty$. Then there exists a unique signed Borel measure $\mu_f$ on $[0,1]^d$ for which 
    \[
    f(x) = \mu_f([0,x]), \quad x \in [0,1]^d.
    \]
    Then we have
    \[
    \text{Var}_{\text{total}} \, \mu_f = V(f) + |f(0)|.
    \]
    Notice that $\text{Var}_{\text{total}} \mu_f$ is the total variation of $\mu_f$ given by the Jordan decomposition of the measure. \\
    Furthermore, if 
    \[
    f(x) = f(0) + f^+(x) - f^-(x)
    \]
    is the Jordan decomposition of $f$, 
    and $\mu_f = \mu_f^+ - \mu_f^-$ is the Jordan decomposition of $\mu_f$, then
    \[
    f^+(x) = \mu_f^+([0, x] \setminus \{0\})
    \quad \text{and} \quad
    f^-(x) = \mu_f^-([0, x] \setminus \{0\}),
    \quad x \in [0,1]^d.
    \]
\end{theorem}
The details of the proof can be found in their paper. We start with the set function $\mu_f^+$ on the elements of all closed axis-parallel boxes contained in $[0,1]^d$ which have one vertex at the origin.
\[
\mu_f^+([0,x]) = f^+(x), \quad \text{for } x \in [0,1]^d. 
\]
We generate the measure on the Borel sigma algebra by the Carathéodory extension theorem, similarly for $\mu_f^-$. The details can be found in Yeh (2006) \citep{yeh2006real} or Bogachev (2007)\citep{bogachev2007measure}. Finally, we define 
\[
\mu_f = \mu_f^+ - \mu_f^- + \delta_{0} f(0).
\]
Then $\mu_f$ is a finite signed Borel measure, and we have 
\[
\mu_f([0,x]) = f^+(x) - f^-(x) + f(0) = f(x).
\]
Thus the identity is rigorously justified,
\[
f(x) = \int_{[0,x]} d\mu_f(u).
\]
For the cube $[0,1]^d$, we define for each subset $s \subset \{1, \ldots, d\}$ the edge $E_s \equiv \{(x(s), 0(-s)) : x \in (0,1]^d\} \subset [0,1]^d$, where $x(s) \equiv (x(j) : j \in s)$ and $0(-s) = (0(j) : j \notin s)$. For the empty subset $s = \emptyset$, we define $E_s = \{0\} \subset [0,1]^d$ as the singleton 0. Note that for $s = \{1, \ldots, d\}$, we have $E_s = (0,1]^d$. For any $x \in [0,1]^d$, consider the $d$-dimensional interval $[0, x]$. We can partition this interval into disjoint "faces" based on which coordinates are strictly positive. Specifically, we have the disjoint union:
\[
[0, x] = \{0\} \cup \bigcup_{\emptyset \neq s \subseteq \{1, \dots, d\}} \left( E_s \cap [0, x] \right)
\]
where $E_s = \{ (u(s), 0(-s)) : u(s) \in (0, 1]^{|s|} \}$. For a given $x$, the intersection $E_s \cap [0, x]$ is non-empty if and only if $x_j > 0$ for all $j \in s$. When non-empty, this set is isometric to the $|s|$-dimensional half-open interval $(0(s), x(s)]$.

Using the identity $f(x) = \mu_f([0, x])$ established above and the finite additivity of the measure $\mu_f$, we obtain:
\[
f(x) = \mu_f(\{0\}) + \sum_{\emptyset \neq s \subseteq \{1, \dots, d\}} \mu_f(E_s \cap [0, x])
\]
By the definition of the measure at the origin, $\mu_f(\{0\}) = f(0)$. Furthermore, for each non-empty $s$, we define the sectional measure $\mu_{f_s}$ as the restriction of $\mu_f$ to the face $E_s$. Identifying $E_s \cap [0, x]$ with the lower-dimensional interval $(0(s), x(s)]$, we can write:
\[
\mu_f(E_s \cap [0, x]) = \int_{(0(s), x(s)]} \mu_{f_s}(du(s))
\]
Summing these contributions yields the desired representation:
\[
f(x) = f(0) + \sum_{s \neq \emptyset, s \subseteq \{1, \dots, d\}} \int_{(0(s), x(s)]} \mu_{f_s}(du(s))
\]
\section{Proof of Theorem~\ref{thm:hal_gram_eigs}}
\label{app:hal-gram-eigs}
\subsection{Eigenstructure of $A_{ij}=\min(i,j)$ via a discrete second-difference problem}

Let $L\in\mathbb{R}^{n\times n}$ be the unit lower-triangular matrix
\[
L_{ij}=\mathbf 1\{i\ge j\}.
\]
A direct calculation gives
\[
A := LL^\top,
\qquad
A_{ij}=\sum_{k=1}^{\min(i,j)}1=\min(i,j).
\]

The inverse of $L$ is the first-difference operator $D:=L^{-1}$ with entries
\[
D_{11}=1,\qquad D_{ii}=1,\quad D_{i,i-1}=-1\ (i\ge2),\qquad 0\ \text{otherwise}.
\]
Hence
\[
A^{-1}=L^{-\top}L^{-1}=D^\top D
=
\begin{bmatrix}
    2 & -1 &  &  &  \\
    -1 & 2 & -1 &  &  \\
     & \ddots & \ddots & \ddots &  \\
     &  & -1 & 2 & -1 \\
     &  &  & -1 & 1
\end{bmatrix}.
\]

Let $x\in\mathbb{R}^n$ and consider the eigenproblem
\[
A^{-1}x=\mu x.
\]
Componentwise, this is
\begin{align*}
 2x_1-x_2 &= \mu x_1,\\
 -x_{r-1}+2x_r-x_{r+1} &= \mu x_r,\qquad r=2,\ldots,n-1,\\
 -x_{n-1}+x_n &= \mu x_n.
\end{align*}
It is convenient to encode the boundary rows using ghost nodes:
define $x_0:=0$ and enforce $x_{n+1}:=x_n$. Then the same interior stencil
\[
-x_{r-1}+2x_r-x_{r+1}=\mu x_r
\]
holds for \emph{all} $r=1,\ldots,n$.\\

Now, we seek solutions of the form $x_r=\sin(r\theta)$. Using
$\sin((r+1)\theta)+\sin((r-1)\theta)=2\cos\theta\,\sin(r\theta)$, the stencil gives
\[
\mu(\theta)=2(1-\cos\theta)=4\sin^2(\theta/2).
\]
The boundary condition $x_{n+1}=x_n$ becomes
\[
\sin((n+1)\theta)=\sin(n\theta)
\iff
2\cos\!\Big(\frac{(2n+1)\theta}{2}\Big)\sin\!\Big(\frac{\theta}{2}\Big)=0.
\]
For nontrivial eigenvectors we take $\sin(\theta/2)\neq 0$, hence
\[
\cos\!\Big(\frac{(2n+1)\theta}{2}\Big)=0
\quad\Rightarrow\quad
\theta_k=\frac{(2k-1)\pi}{2n+1},\qquad k=1,\ldots,n.
\]
Thus an eigenbasis of $A^{-1}$ is given by
\[
x^{(k)}_r=\sin(r\theta_k),
\qquad
\mu_k=2\bigl(1-\cos\theta_k\bigr).
\]

\paragraph{Step 4: Convert to eigenpairs of $A$ and normalize.}
Since $A^{-1}x^{(k)}=\mu_k x^{(k)}$, the eigenvalues of $A$ are
\[
\lambda_k(A)=\frac{1}{\mu_k}
=\frac{1}{2\bigl(1-\cos\theta_k\bigr)}
=\frac{1}{4\sin^2(\theta_k/2)}.
\]
Moreover,
\(
\sum_{r=1}^n \sin^2(r\theta_k) = \frac{2n+1}{4},
\)
so the normalized eigenvectors are
\[
u_k(r)=\sqrt{\frac{4}{2n+1}}\sin(r\theta_k),
\qquad r=1,\ldots,n.
\]
Therefore $A=U\,\mathrm{diag}(\lambda_1(A),\ldots,\lambda_n(A))\,U^\top$ with
$U=[u_1\ \cdots\ u_n$].

\subsection{Continuum limit and Fourier/Sturm--Liouville connection}

Let $h:=\frac{1}{n+1}$ and $s_i:=ih$ for $i=1,\ldots,n$.  Then
\[
A_{ij}=\min(i,j)=\frac{1}{h}\min(s_i,s_j),
\qquad\text{equivalently}\qquad
\min(s_i,s_j)=h\,A_{ij}.
\]
Define the continuum integral operator
\[
(Tf)(s)=\int_0^1 \min(s,t)\,f(t)\,dt,\qquad s\in[0,1].
\]
For a smooth $f$ and the grid vector $f_i:=f(s_i)$, the Riemann-sum approximation gives
\[
(Tf)(s_i)\;\approx\; h\sum_{j=1}^n \min(s_i,s_j)\,f(s_j)
\;=\; h\sum_{j=1}^n (hA_{ij})\,f_j
\;=\; h^2 (Af)_i.
\]
Thus the \emph{scaled matrix} $h^2A$ is the natural discretization of $T$, and the
eigenvalues satisfy $h^2\lambda_k(A)\to \lambda_k(T)$ as $n\to\infty$ (for fixed $k$).

\paragraph{Green's function and Sturm--Liouville operator.}
The kernel $K_\infty(s,t)=\min(s,t)$ is the Green's function of the
self-adjoint operator $-\frac{d^2}{ds^2}$ on $[0,1]$ with mixed boundary conditions
\[
u(0)=0,\qquad u'(1)=0.
\]
Accordingly, the associated Sturm--Liouville eigenproblem is
\begin{equation}\label{eq:SL}
-\,\phi''(s)=\mu\,\phi(s)\quad (0<s<1),\qquad \phi(0)=0,\ \ \phi'(1)=0.
\end{equation}
Its eigenpairs are the half-shifted sines
\[
\phi_k(s)=\sqrt{2}\,\sin\!\big((k+\tfrac12)\pi s\big),\qquad
\mu_k=((k+\tfrac12)\pi)^2,\qquad k=0,1,2,\ldots,
\]
and since $T$ is the inverse of $-\frac{d^2}{ds^2}$ under these boundary conditions,
\[
T\phi_k=\lambda_k(T)\phi_k,\qquad \lambda_k(T)=\frac{1}{\mu_k}
=\frac{1}{((k+\tfrac12)\pi)^2}.
\]

\paragraph{Matching to the discrete eigensystem.}
Recall the discrete eigenpairs of $A$:
\[
\theta_k=\frac{(2k-1)\pi}{2n+1},\qquad
u_k(i)=\sqrt{\frac{4}{2n+1}}\sin(i\theta_k),\qquad
\lambda_k(A)=\frac{1}{2(1-\cos\theta_k)},\qquad k=1,\ldots,n.
\]
For fixed $k$ and $n\to\infty$, we have $\theta_k\sim (k-\tfrac12)\pi h$ and hence
\[
h^2\lambda_k(A)
=\frac{h^2}{2(1-\cos\theta_k)}
\longrightarrow
\frac{1}{((k-\tfrac12)\pi)^2}
=\lambda_{k-1}(T),
\]
while $u_k(i)\approx \phi_{k-1}(s_i)$.  This is exactly the Karhunen--Lo\`eve
eigendecomposition of Brownian motion on $[0,1]$ with covariance kernel $\min(s,t)$.

\section{Truncated HA kernel}\label{sec:truncha}

Expanding the definition of $H_{\le m}$ gives
\begin{align*}
K_{\le m}(x,x')
  &= \sum_{i=1}^n \ \sum_{\substack{s \subseteq [d] \\ 1 \le |s| \le m}}
     \Bigg(\prod_{j\in s} 1\{X_{i,j} \le x_j\}\Bigg)
     \Bigg(\prod_{j\in s} 1\{X_{i,j} \le x'_j\}\Bigg) \\
  &\overset{(a)}{=} \sum_{i=1}^n \ \sum_{\substack{s \subseteq [d] \\ 1 \le |s| \le m}}
     \prod_{j\in s} 1\{X_{i,j} \le (x \wedge x')_j\},
\end{align*}
where (a) follows from the identity
\[
1\{X_{i,j} \le x_j\}\,1\{X_{i,j} \le x'_j\}
    = 1\{X_{i,j} \le \min(x_j,x'_j)\}.
\]

For each $i$, let
\[
s_i(x,x') := \{\, j \in [d] : X_{i,j} \le (x \wedge x')_j \,\},
\qquad
t_i(x,x') := |s_i(x,x')|.
\]
The product $\prod_{j\in s} 1\{X_{i,j} \le (x \wedge x')_j\}$ equals $1$ if and
only if $s \subseteq s_i(x,x')$, and equals $0$ otherwise. Thus the inner sum
counts all subsets of $s_i(x,x')$ of size at most $m$:
\begin{align*}
K_{\le m}(x,x')
  &= \sum_{i=1}^n \ \sum_{\substack{s \subseteq s_i(x,x') \\ 1 \le |s| \le m}} 1
   \;=\; \sum_{i=1}^n \ \sum_{r=1}^{\min\{m,\,t_i(x,x')\}} \binom{t_i(x,x')}{r}.
\end{align*}

\section{Proof of Theorem \ref{thm:PCHA}}

For (i), note that
\begin{equation}\label{eq:ZtZ}
    Z_k^{\top} Z_k
    \;=\; D_{k}^{1/2} U_{k}^{\top} U_{k} D_{k}^{1/2}
    \;=\; D_{k}
\end{equation} By expanding the objective:
\[
\frac{1}{2n}\|Y-Z_k\beta\|_2^{2}+\frac{\lambda}{2}\|\beta\|_2^2
= \frac{1}{2n}\Bigl(Y^{\top}Y -2\beta^{\top}Z_k^{\top}Y + \beta^{\top}Z_k^{\top}Z_k\beta\Bigr)
+ \frac{\lambda}{2}\beta^{\top}\beta.
\]
Substituting \eqref{eq:ZtZ}, the first‐order condition is
\[
-Z_k^{\top}Y + D_{k}\beta +n \lambda\beta \;=\; 0
\quad\Longrightarrow\quad
(D_{k}+n \lambda I_k)\beta = Z_k^{\top}Y.
\]
Hence
\(
\hat\beta^{\mathrm{PCHAR}}_{k,\lambda}
= (D_{k}+n \lambda I_k)^{-1}Z_k^{\top}Y
= (D_{k}+n \lambda I_k)^{-1}D_{k}^{1/2}U_{k}^{\top}Y,
\)
and $\hat f = Z_k \hat\beta$ yields the fitted‐value expression.

For (ii), using \eqref{eq:ZtZ}, the objective separates as
\[
\frac{1}{2n}\|Y-Z_k\beta\|_2^{2}+\lambda\|\beta\|_1
= \frac{1}{2n}Y^\top Y
+ \sum_{j=1}^k \left\{\frac{d_j}{2n}\beta_j^2-\frac{w_j\beta_j}{n}+\lambda |\beta_j|\right\},
\]
where $w_j\coloneqq(Z_k^{\top}Y)_j$. For each $j$, we minimize
\[
q_j(\beta_j):=\frac{d_j}{2n}\beta_j^2 - \frac{w_j\beta_j}{n}+ \lambda|\beta_j|.
\]
By the subgradient KKT condition, the unique minimizer for $d_j>0$ is the
soft-threshold of $w_j/d_j$ with threshold $n\lambda/d_j$:
\[
\beta_j^* = \mathsf{S}\!\left(\frac{w_j}{d_j},\frac{n\lambda}{d_j}\right)
= \frac{1}{d_j}\,\mathsf{S}(w_j,n\lambda)
= \frac{1}{d_j}\,\mathrm{sign}(w_j)\bigl(|w_j|-n\lambda\bigr)_+.
\]
For coordinates with $d_j=0$, the fitted value does not depend on $\beta_j$, so we set
$\beta_j=0$. Stacking the coordinates yields
\[
\hat\beta^{\mathrm{PCHAL}}_{k,\lambda}
=
D_k^{-1}\mathrm{sign}(Z_k^\top Y)
\left(|Z_k^\top Y|-n\lambda\right)_+,
\]
which proves the second claim.

\section{Centering the HA design}
\label{sec:center-ha-design}

Principal component analysis is fundamentally applied to centered data. In our
setting, the design matrix \(H \in \mathbb{R}^{n\times p}\) contains step–function
basis evaluations that are generally non–zero–mean. If PCA is applied directly
to \(H\), the first principal component will be dominated by the mean level of
the basis dictionary rather than by variation around the mean. To remove this
artificial location effect, it is standard to \emph{center} the columns of \(H\):
\[
\tilde H \;:=\; H - \mathbf{1}_n\mu^\top,
\qquad
\mu := \frac{1}{n} H^\top \mathbf{1}_n,
\]
where \(\mathbf{1}_n \in \mathbb{R}^n\) denotes the vector of ones. Thus,
\(\mu_j\) is the empirical mean of the \(j\)-th basis function across the sample,
and every column of \(\tilde H\) has mean zero. Define the centering matrix
\[
J := I_n - \frac{1}{n}\mathbf{1}_n\mathbf{1}_n^\top.
\]
A direct computation shows
\[
\tilde H = J H,
\qquad\text{and}\qquad
\tilde H \tilde H^\top = J H H^\top J.
\]
Therefore, centering the columns of \(H\) corresponds to double–centering the kernel
matrix \(K = H H^\top\):
\[
\tilde K := \tilde H \tilde H^\top
= J K J.
\]
This identity will be used repeatedly when constructing the PC-HA embedding,
allowing us to diagonalize \(\tilde H\tilde H^\top\) without explicitly forming the
centered design matrix \(\tilde H\).

We now show how centering forces the left singular vectors to have mean zero.
Suppose \(\tilde H\) admits the singular value decomposition
\[
\tilde H = U \sqrt{D} V^\top,
\]
where \(U \in \mathbb{R}^{n\times n}\) has orthonormal columns and \(\sqrt{D}\) is diagonal
with nonnegative entries. We have
\[
\tilde H^\top \mathbf{1}_n = 0
\;\;\Longrightarrow\;\;
V \sqrt{D}\, U^\top \mathbf{1}_n = 0.
\]
On the subspace corresponding to nonzero singular values, \(\sqrt{D}\) and \(V\) are
invertible, so this implies
\[
U^\top \mathbf{1}_n = 0
\]
on that subspace. Hence each left singular vector associated with a nonzero
singular value has empirical mean zero. Consequently, every column of the PC
score matrix
\[
Z_k = U_k D_k^{1/2}
\]
also has mean zero, since it is a scalar multiple of a mean-zero column of \(U_k\).

To make the role of centering explicit, consider the standard univariate linear
regression model (modulo residuals for ease of notation) with an intercept:
\[
Y_i = \beta_0 + \beta_1 Z_i,\qquad i=1,\ldots,n.
\]
Assume that the predictor has already been centered, i.e.\ its empirical mean is
zero:
\[
\frac{1}{n}\sum_{i=1}^n Z_i = 0.
\]
Let \(\bar Y := \frac{1}{n}\sum_{i=1}^n Y_i\) denote the empirical mean of the
response, and define the centered response
\[
\tilde Y_i := Y_i - \bar Y,\qquad i=1,\ldots,n.
\]

Taking the empirical mean of both sides of the model gives
\[
\bar Y = \beta_0 + \beta_1 \cdot \underbrace{\frac{1}{n}\sum_{i=1}^n Z_i}_{=\,0}
= \beta_0.
\]
Thus, the intercept is precisely the mean of \(Y\):
\[
\beta_0 = \bar Y.
\]

Substituting \(Y_i = \tilde Y_i + \bar Y\) into the original model,
\[
\tilde Y_i + \bar Y = \beta_0 + \beta_1 Z_i.
\]
Since \(\beta_0 = \bar Y\), these terms cancel, leaving
\[
\tilde Y_i = \beta_1 Z_i.
\]
Hence the centered model is
\[
\tilde Y = \beta_1 Z,
\]
with no intercept term.  The effect of centering is therefore to absorb
\(\beta_0\) into the mean of \(Y\).

Importantly, this shows that when a predictor has mean zero, estimating
\(\beta_1\) by regressing \(\tilde Y\) on \(Z\) \emph{without} an intercept yields the
same slope coefficient as the original regression of \(Y\) on \(Z\) \emph{with} an
intercept.  If one wishes to recover the intercept afterward, it is simply
\[
\hat\beta_0 = \bar Y.
\]

This univariate example illustrates the general multivariate principle used in
our construction: once the design matrix is column–centered, the intercept in
the linear model is fully determined by the mean of the response and does not
need to be explicitly included in the regression. This simplification is
particularly convenient when working with principal components, since the PC
score matrix \(Z_k = U_k D_k^{1/2}\) inherits the mean–zero property of \(U_k\).

\section{Selecting the Maximum Interaction Order}
\label{appendix:interaction-order-cv}

The covariates are generated independently as
\[
X_{ij} \sim \mathrm{Unif}(-1,1), \qquad i=1,\ldots,n,\;\; j=1,\ldots,d,
\]
with $d=3$ throughout. The true regression function contains main effects and second-order interactions:
\[
g(X) \;=\; \beta^\top X \;+\; 0.3\big(X_1X_2 \;-\; 1.5 X_2X_3\big),
\]
where $\beta = (1.2,-1.0,0.8)$, and we generate
\[
Y = g(X) + \varepsilon, \qquad \varepsilon \sim N(0,0.03^2).
\]
Thus the true model corresponds to maximum interaction order $m_0=2$. We consider the candidate library
\[
\mathcal{F} = \{\hat{f}_{m,\lambda} : m \in \{1,2,3\},\, \lambda \in \Lambda\},
\]
where $\Lambda$ is a fixed grid. For each training sample size $n \in \{100,300,500\}$, and each $m \in \{1,2,3\}$, we compute the cross-validated risk curve
\[
\lambda \;\mapsto\; \widehat{R}_{\mathrm{CV}}(m,\lambda),
\]
and record the minimum (profiled) risk
\[
\widehat{R}_{\mathrm{CV}}(m) \;=\; \min_{\lambda \in \Lambda} \widehat{R}_{\mathrm{CV}}(m,\lambda).
\]
The selected interaction degree is then
\[
\hat{m} = \arg\min_{m \in \{1,2,3\}} \widehat{R}_{\mathrm{CV}}(m).
\]

To evaluate oracle performance, we independently generate a fresh test sample of size $n_{\mathrm{test}}=5000$ and compute the true prediction risk for each $(m,\lambda)$. The oracle interaction degree is defined as
\[
m^* = \arg\min_{m \in \{1,2,3\}} \min_{\lambda \in \Lambda} R_{\mathrm{test}}(m,\lambda).
\]

\medskip

Tables~\ref{tab:oracle-props} and~\ref{tab:cv-props} report, over repeated simulations, the empirical frequencies with which cross-validation selects each $m$, along with the oracle selection frequencies. The oracle inequality essentially states that these two tables should be similar, in the sense that their behavior differs by a term of order $\mathcal{O}(\frac{1}{n})$. Table~\ref{tab:oracle-props} illustrates the behavior of the oracle selector, which has access to the full data-generating distribution and can therefore evaluate and minimize the true MSE. As the training sample size increases, the oracle selects models of increasing complexity, since higher interaction orders can be included without incurring overfitting. Table~\ref{tab:cv-props} shows the corresponding behavior of the cross-validation (CV) selector, which, as established by the oracle inequality, behaves like the oracle up to an $\mathcal{O}(\frac{1}{n})$ term. Notice that, for larger sample sizes, the oracle no longer selects interaction order 1, since it is not sufficiently expressive for the ground truth. However, for small sample sizes, $m=1$ is selected most of the time, since the oracle deems higher-order interactions overly complex for the available data. The CV selector exhibits the same behavior. 

The results illustrate the finite-sample adaptivity behavior predicted by the oracle inequality. When $n=100$, the signal in the interaction terms is weak relative to sampling variability, and both the oracle and cross-validation frequently prefer $m=1$, i.e., no interactions. At $n=300$, the signal-to-noise ratio improves, and both selectors exhibit mixed behavior, allocating non-negligible mass to $m=2$. Finally, at $n=500$, the oracle clearly favors $m=2$, and cross-validation also selects $m=2$ with high frequency. In other words, as the sample size increases, cross-validation smoothly tracks the oracle transition from preferring simpler main-effects models to preferring interaction models once such structure becomes statistically identifiable.

This matches theoretical arguments: since the candidate library size is fixed, the cross-validated selector performs within $O_p(n^{-1})$ of the oracle risk, and therefore identifies the correct interaction degree once the problem is sufficiently well-conditioned at the available sample size.

Note that we are not working in a fully correctly specified finite-dimensional parametric model, where the regression function is exactly a linear combination of products of indicator functions so that the true active set would be contained in our library. In such a setting, the oracle estimator is simply the correctly specified parametric model, and its estimation error converges at the usual parametric rate $n^{-1/2}$. The cross-validation selector would then recover this model, and both the oracle and cross-validated risks would converge at a parametric rate. In our case, however, we are approximating the true regression surface, so the oracle itself depends on the sample size $n$ through the bias--variance tradeoff. Cross-validation tracks this evolving oracle, rather than converging to a fixed finite-dimensional true model.

\begin{table}[t]
	\centering
	\begin{minipage}{0.48\textwidth}
		\centering
		\begin{tabular}{lccc}
			\toprule
			$n$&  $m=1$ & $m=2$ &  $m=3$ \\
			\midrule
			$100$ & 0.61 & 0.24 & 0.15 \\
			$300$ & 0.00 & 0.54 & 0.46 \\
			$500$ & 0.00 & 0.62 & 0.38 \\
			\bottomrule
		\end{tabular}
		\caption{Proportion of runs in which each $m$ minimized the {oracle} MSE.}
		\label{tab:oracle-props}
	\end{minipage}
	\hfill
	\begin{minipage}{0.48\textwidth}
		\centering
		\begin{tabular}{lccc}
			\toprule
			$n$ &  $m=1$ &  $m=2$ &  $m=3$ \\
			\midrule
			$100$ & 0.77 & 0.18 & 0.05 \\
			$300$ & 0.05 & 0.51 & 0.44 \\
			$500$ & 0.00 & 0.64 & 0.36 \\
			\bottomrule
		\end{tabular}
		\caption{Proportion of runs in which each $m$ minimized the {cross-validated} MSE.}
		\label{tab:cv-props}
	\end{minipage}
\end{table}

\bibliographystyle{apacite}
\bibliography{bib}
\end{document}